\documentclass[10pt,twocolumn,letterpaper]{article}

\usepackage[pagenumbers]{cvpr} 

\usepackage{graphicx}
\usepackage{amsmath}
\usepackage{amssymb}
\usepackage{booktabs}
\usepackage{amsfonts,bm}
\usepackage[page]{appendix} 

\usepackage[ruled,vlined,linesnumbered,algo2e]{algorithm2e}
  \SetKwInput{KwIn}{Inputs}
  \SetKwInput{KwNetwork}{Network}
  \SetKwInput{KwHp}{Hyper-parameters}
  \SetKwFunction{Ffor}{for}
  \SetKwProg{Fn}{}{:}{}
\SetKwInput{KwReturn}{return}
  \DontPrintSemicolon
  
\usepackage{xcolor}

\SetCommentSty{mycommfont}

\usepackage[pagebackref,breaklinks,colorlinks]{hyperref}

\usepackage[capitalize]{cleveref}
\crefname{section}{Sec.}{Secs.}
\Crefname{section}{Section}{Sections}
\Crefname{table}{Table}{Tables}
\crefname{table}{Tab.}{Tabs.}

\def\cvprPaperID{*****} 
\def\confName{CVPR}
\def\confYear{2023}

\begin{document}
\title{Improving Robust Generalization \\ by Direct PAC-Bayesian Bound Minimization}

\author{
Zifan Wang$^1$\thanks{Work done during internship in Google.} \quad Nan Ding$^2$\thanks{Corresponding author}\quad Tomer Levinboim$^2$ \quad Xi Chen$^2$  \quad Radu Soricut$^2$ \\
Carnegie Mellon University$^1$  \quad Google Research$^2$   \\
{\tt\small zifan@cmu.edu, \{dingnan, tomerl, chillxichen, rsoricut\}@google.com}
}

\maketitle
\newcommand{\todo}[1]{\textcolor{blue}{(TODO): #1}}
\newcommand{\zw}[1]{\textcolor{red}{ZW: #1}}
\newcommand{\nd}[1]{\textcolor{orange}{ND: #1}}
\newcommand{\tl}[1]{\textcolor{green}{TL: #1}}
\newcommand{\xc}[1]{\textcolor{pink}{XC: #1}}

\newcommand{\defeq}{\overset{\text{\tiny def}}{=}}
\newcommand{\jacobian}{\triangledown}
\newcommand{\hessian}{\triangledown^2}

\newcommand{\trainloss}[1]{\hat{L}(#1, D^m)}
\newcommand{\testloss}[1]{L(#1, \mathcal{D})}
\newcommand{\robusttrainloss}[1]{\hat{R}(#1, D^m)}
\newcommand{\robusttrainlosss}{\hat{R}}
\newcommand{\robusttestloss}[1]{R(#1, \mathcal{D})}
\newcommand{\atloss}[1]{\hat{R}_{\texttt{AT}}(#1, D^m)}
\newcommand{\atlossn}{\hat{R}_{\texttt{AT}}}
\newcommand{\tradesloss}[1]{\hat{R}_{\texttt{T}}(#1, D^m)}
\newcommand{\rloss}[1]{\hat{R}_{\texttt{*}}(#1, D^m)}

\newcommand{\base}{\texttt{base}}
\newcommand{\AWP}{\texttt{AWP}}
\newcommand{\AWPT}{\texttt{AWPT}}
\newcommand{\WA}{\texttt{SWA}}
\newcommand{\STO}{\texttt{S2O}}
\newcommand{\se}[1]{\footnotesize{\pm #1}}
\newcommand{\softmaxf}[1]{s({#1})}
\newtheorem{definition}{Definition}
\newtheorem{theorem}{Theorem}
\newtheorem{lemma}{Lemma}
\newtheorem{proposition}{Proposition}
\newtheorem{remark}{Remark}
\newtheorem{example}{Example}


\newcommand{\figleft}{{\em (Left)}}
\newcommand{\figcenter}{{\em (Center)}}
\newcommand{\figright}{{\em (Right)}}
\newcommand{\figtop}{{\em (Top)}}
\newcommand{\figbottom}{{\em (Bottom)}}
\newcommand{\captiona}{{\em (a)}}
\newcommand{\captionb}{{\em (b)}}
\newcommand{\captionc}{{\em (c)}}
\newcommand{\captiond}{{\em (d)}}

\newcommand{\newterm}[1]{{\bf #1}}

\def\figref#1{figure~\ref{#1}}
\def\Figref#1{Figure~\ref{#1}}
\def\twofigref#1#2{figures \ref{#1} and \ref{#2}}
\def\quadfigref#1#2#3#4{figures \ref{#1}, \ref{#2}, \ref{#3} and \ref{#4}}
\def\secref#1{section~\ref{#1}}
\def\Secref#1{Section~\ref{#1}}
\def\twosecrefs#1#2{sections \ref{#1} and \ref{#2}}
\def\secrefs#1#2#3{sections \ref{#1}, \ref{#2} and \ref{#3}}
\def\eqref#1{equation~\ref{#1}}
\def\Eqref#1{Equation~\ref{#1}}
\def\plaineqref#1{\ref{#1}}
\def\chapref#1{chapter~\ref{#1}}
\def\Chapref#1{Chapter~\ref{#1}}
\def\rangechapref#1#2{chapters\ref{#1}--\ref{#2}}
\def\algref#1{algorithm~\ref{#1}}
\def\Algref#1{Algorithm~\ref{#1}}
\def\twoalgref#1#2{algorithms \ref{#1} and \ref{#2}}
\def\Twoalgref#1#2{Algorithms \ref{#1} and \ref{#2}}
\def\partref#1{part~\ref{#1}}
\def\Partref#1{Part~\ref{#1}}
\def\twopartref#1#2{parts \ref{#1} and \ref{#2}}

\def\ceil#1{\lceil #1 \rceil}
\def\floor#1{\lfloor #1 \rfloor}
\def\1{\bm{1}}
\newcommand{\train}{\mathcal{D}}
\newcommand{\valid}{\mathcal{D_{\mathrm{valid}}}}
\newcommand{\test}{\mathcal{D_{\mathrm{test}}}}

\def\eps{{\epsilon}}

\def\reta{{\textnormal{$\eta$}}}
\def\ra{{\textnormal{a}}}
\def\rb{{\textnormal{b}}}
\def\rc{{\textnormal{c}}}
\def\rd{{\textnormal{d}}}
\def\re{{\textnormal{e}}}
\def\rf{{\textnormal{f}}}
\def\rg{{\textnormal{g}}}
\def\rh{{\textnormal{h}}}
\def\ri{{\textnormal{i}}}
\def\rj{{\textnormal{j}}}
\def\rk{{\textnormal{k}}}
\def\rl{{\textnormal{l}}}
\def\rn{{\textnormal{n}}}
\def\ro{{\textnormal{o}}}
\def\rp{{\textnormal{p}}}
\def\rq{{\textnormal{q}}}
\def\rr{{\textnormal{r}}}
\def\rs{{\textnormal{s}}}
\def\rt{{\textnormal{t}}}
\def\ru{{\textnormal{u}}}
\def\rv{{\textnormal{v}}}
\def\rw{{\textnormal{w}}}
\def\rx{{\textnormal{x}}}
\def\ry{{\textnormal{y}}}
\def\rz{{\textnormal{z}}}

\def\rvepsilon{{\mathbf{\epsilon}}}
\def\rvtheta{{\mathbf{\theta}}}
\def\rva{{\mathbf{a}}}
\def\rvb{{\mathbf{b}}}
\def\rvc{{\mathbf{c}}}
\def\rvd{{\mathbf{d}}}
\def\rve{{\mathbf{e}}}
\def\rvf{{\mathbf{f}}}
\def\rvg{{\mathbf{g}}}
\def\rvh{{\mathbf{h}}}
\def\rvu{{\mathbf{i}}}
\def\rvj{{\mathbf{j}}}
\def\rvk{{\mathbf{k}}}
\def\rvl{{\mathbf{l}}}
\def\rvm{{\mathbf{m}}}
\def\rvn{{\mathbf{n}}}
\def\rvo{{\mathbf{o}}}
\def\rvp{{\mathbf{p}}}
\def\rvq{{\mathbf{q}}}
\def\rvr{{\mathbf{r}}}
\def\rvs{{\mathbf{s}}}
\def\rvt{{\mathbf{t}}}
\def\rvu{{\mathbf{u}}}
\def\rvv{{\mathbf{v}}}
\def\rvw{{\mathbf{w}}}
\def\rvx{{\mathbf{x}}}
\def\rvy{{\mathbf{y}}}
\def\rvz{{\mathbf{z}}}

\def\erva{{\textnormal{a}}}
\def\ervb{{\textnormal{b}}}
\def\ervc{{\textnormal{c}}}
\def\ervd{{\textnormal{d}}}
\def\erve{{\textnormal{e}}}
\def\ervf{{\textnormal{f}}}
\def\ervg{{\textnormal{g}}}
\def\ervh{{\textnormal{h}}}
\def\ervi{{\textnormal{i}}}
\def\ervj{{\textnormal{j}}}
\def\ervk{{\textnormal{k}}}
\def\ervl{{\textnormal{l}}}
\def\ervm{{\textnormal{m}}}
\def\ervn{{\textnormal{n}}}
\def\ervo{{\textnormal{o}}}
\def\ervp{{\textnormal{p}}}
\def\ervq{{\textnormal{q}}}
\def\ervr{{\textnormal{r}}}
\def\ervs{{\textnormal{s}}}
\def\ervt{{\textnormal{t}}}
\def\ervu{{\textnormal{u}}}
\def\ervv{{\textnormal{v}}}
\def\ervw{{\textnormal{w}}}
\def\ervx{{\textnormal{x}}}
\def\ervy{{\textnormal{y}}}
\def\ervz{{\textnormal{z}}}

\def\rmA{{\mathbf{A}}}
\def\rmB{{\mathbf{B}}}
\def\rmC{{\mathbf{C}}}
\def\rmD{{\mathbf{D}}}
\def\rmE{{\mathbf{E}}}
\def\rmF{{\mathbf{F}}}
\def\rmG{{\mathbf{G}}}
\def\rmH{{\mathbf{H}}}
\def\rmI{{\mathbf{I}}}
\def\rmJ{{\mathbf{J}}}
\def\rmK{{\mathbf{K}}}
\def\rmL{{\mathbf{L}}}
\def\rmM{{\mathbf{M}}}
\def\rmN{{\mathbf{N}}}
\def\rmO{{\mathbf{O}}}
\def\rmP{{\mathbf{P}}}
\def\rmQ{{\mathbf{Q}}}
\def\rmR{{\mathbf{R}}}
\def\rmS{{\mathbf{S}}}
\def\rmT{{\mathbf{T}}}
\def\rmU{{\mathbf{U}}}
\def\rmV{{\mathbf{V}}}
\def\rmW{{\mathbf{W}}}
\def\rmX{{\mathbf{X}}}
\def\rmY{{\mathbf{Y}}}
\def\rmZ{{\mathbf{Z}}}

\def\ermA{{\textnormal{A}}}
\def\ermB{{\textnormal{B}}}
\def\ermC{{\textnormal{C}}}
\def\ermD{{\textnormal{D}}}
\def\ermE{{\textnormal{E}}}
\def\ermF{{\textnormal{F}}}
\def\ermG{{\textnormal{G}}}
\def\ermH{{\textnormal{H}}}
\def\ermI{{\textnormal{I}}}
\def\ermJ{{\textnormal{J}}}
\def\ermK{{\textnormal{K}}}
\def\ermL{{\textnormal{L}}}
\def\ermM{{\textnormal{M}}}
\def\ermN{{\textnormal{N}}}
\def\ermO{{\textnormal{O}}}
\def\ermP{{\textnormal{P}}}
\def\ermQ{{\textnormal{Q}}}
\def\ermR{{\textnormal{R}}}
\def\ermS{{\textnormal{S}}}
\def\ermT{{\textnormal{T}}}
\def\ermU{{\textnormal{U}}}
\def\ermV{{\textnormal{V}}}
\def\ermW{{\textnormal{W}}}
\def\ermX{{\textnormal{X}}}
\def\ermY{{\textnormal{Y}}}
\def\ermZ{{\textnormal{Z}}}

\def\vzero{{\bm{0}}}
\def\vone{{\bm{1}}}
\def\vmu{{\bm{\mu}}}
\def\vtheta{{\bm{\theta}}}
\def\va{{\bm{a}}}
\def\vb{{\bm{b}}}
\def\vc{{\bm{c}}}
\def\vd{{\bm{d}}}
\def\ve{{\bm{e}}}
\def\vf{{\bm{f}}}
\def\vg{{\bm{g}}}
\def\vh{{\bm{h}}}
\def\vi{{\bm{i}}}
\def\vj{{\bm{j}}}
\def\vk{{\bm{k}}}
\def\vl{{\bm{l}}}
\def\vm{{\bm{m}}}
\def\vn{{\bm{n}}}
\def\vo{{\bm{o}}}
\def\vp{{\bm{p}}}
\def\vq{{\bm{q}}}
\def\vr{{\bm{r}}}
\def\vs{{\bm{s}}}
\def\vt{{\bm{t}}}
\def\vu{{\bm{u}}}
\def\vv{{\bm{v}}}
\def\vw{{\bm{w}}}
\def\vx{{\bm{x}}}
\def\vy{{\bm{y}}}
\def\vz{{\bm{z}}}

\def\evalpha{{\alpha}}
\def\evbeta{{\beta}}
\def\evepsilon{{\epsilon}}
\def\evlambda{{\lambda}}
\def\evomega{{\omega}}
\def\evmu{{\mu}}
\def\evpsi{{\psi}}
\def\evsigma{{\sigma}}
\def\evtheta{{\theta}}
\def\eva{{a}}
\def\evb{{b}}
\def\evc{{c}}
\def\evd{{d}}
\def\eve{{e}}
\def\evf{{f}}
\def\evg{{g}}
\def\evh{{h}}
\def\evi{{i}}
\def\evj{{j}}
\def\evk{{k}}
\def\evl{{l}}
\def\evm{{m}}
\def\evn{{n}}
\def\evo{{o}}
\def\evp{{p}}
\def\evq{{q}}
\def\evr{{r}}
\def\evs{{s}}
\def\evt{{t}}
\def\evu{{u}}
\def\evv{{v}}
\def\evw{{w}}
\def\evx{{x}}
\def\evy{{y}}
\def\evz{{z}}

\def\mA{{\bm{A}}}
\def\mB{{\bm{B}}}
\def\mC{{\bm{C}}}
\def\mD{{\bm{D}}}
\def\mE{{\bm{E}}}
\def\mF{{\bm{F}}}
\def\mG{{\bm{G}}}
\def\mH{{\bm{H}}}
\def\mI{{\bm{I}}}
\def\mJ{{\bm{J}}}
\def\mK{{\bm{K}}}
\def\mL{{\bm{L}}}
\def\mM{{\bm{M}}}
\def\mN{{\bm{N}}}
\def\mO{{\bm{O}}}
\def\mP{{\bm{P}}}
\def\mQ{{\bm{Q}}}
\def\mR{{\bm{R}}}
\def\mS{{\bm{S}}}
\def\mT{{\bm{T}}}
\def\mU{{\bm{U}}}
\def\mV{{\bm{V}}}
\def\mW{{\bm{W}}}
\def\mX{{\bm{X}}}
\def\mY{{\bm{Y}}}
\def\mZ{{\bm{Z}}}
\def\mBeta{{\bm{\beta}}}
\def\mPhi{{\bm{\Phi}}}
\def\mLambda{{\bm{\Lambda}}}
\def\mSigma{{\bm{\Sigma}}}

\newcommand{\tens}[1]{\bm{\mathsfit{#1}}}
\def\tA{{\tens{A}}}
\def\tB{{\tens{B}}}
\def\tC{{\tens{C}}}
\def\tD{{\tens{D}}}
\def\tE{{\tens{E}}}
\def\tF{{\tens{F}}}
\def\tG{{\tens{G}}}
\def\tH{{\tens{H}}}
\def\tI{{\tens{I}}}
\def\tJ{{\tens{J}}}
\def\tK{{\tens{K}}}
\def\tL{{\tens{L}}}
\def\tM{{\tens{M}}}
\def\tN{{\tens{N}}}
\def\tO{{\tens{O}}}
\def\tP{{\tens{P}}}
\def\tQ{{\tens{Q}}}
\def\tR{{\tens{R}}}
\def\tS{{\tens{S}}}
\def\tT{{\tens{T}}}
\def\tU{{\tens{U}}}
\def\tV{{\tens{V}}}
\def\tW{{\tens{W}}}
\def\tX{{\tens{X}}}
\def\tY{{\tens{Y}}}
\def\tZ{{\tens{Z}}}

\def\gA{{\mathcal{A}}}
\def\gB{{\mathcal{B}}}
\def\gC{{\mathcal{C}}}
\def\gD{{\mathcal{D}}}
\def\gE{{\mathcal{E}}}
\def\gF{{\mathcal{F}}}
\def\gG{{\mathcal{G}}}
\def\gH{{\mathcal{H}}}
\def\gI{{\mathcal{I}}}
\def\gJ{{\mathcal{J}}}
\def\gK{{\mathcal{K}}}
\def\gL{{\mathcal{L}}}
\def\gM{{\mathcal{M}}}
\def\gN{{\mathcal{N}}}
\def\gO{{\mathcal{O}}}
\def\gP{{\mathcal{P}}}
\def\gQ{{\mathcal{Q}}}
\def\gR{{\mathcal{R}}}
\def\gS{{\mathcal{S}}}
\def\gT{{\mathcal{T}}}
\def\gU{{\mathcal{U}}}
\def\gV{{\mathcal{V}}}
\def\gW{{\mathcal{W}}}
\def\gX{{\mathcal{X}}}
\def\gY{{\mathcal{Y}}}
\def\gZ{{\mathcal{Z}}}

\def\sA{{\mathbb{A}}}
\def\sB{{\mathbb{B}}}
\def\sC{{\mathbb{C}}}
\def\sD{{\mathbb{D}}}
\def\sF{{\mathbb{F}}}
\def\sG{{\mathbb{G}}}
\def\sH{{\mathbb{H}}}
\def\sI{{\mathbb{I}}}
\def\sJ{{\mathbb{J}}}
\def\sK{{\mathbb{K}}}
\def\sL{{\mathbb{L}}}
\def\sM{{\mathbb{M}}}
\def\sN{{\mathbb{N}}}
\def\sO{{\mathbb{O}}}
\def\sP{{\mathbb{P}}}
\def\sQ{{\mathbb{Q}}}
\def\sR{{\mathbb{R}}}
\def\sS{{\mathbb{S}}}
\def\sT{{\mathbb{T}}}
\def\sU{{\mathbb{U}}}
\def\sV{{\mathbb{V}}}
\def\sW{{\mathbb{W}}}
\def\sX{{\mathbb{X}}}
\def\sY{{\mathbb{Y}}}
\def\sZ{{\mathbb{Z}}}

\def\emLambda{{\Lambda}}
\def\emA{{A}}
\def\emB{{B}}
\def\emC{{C}}
\def\emD{{D}}
\def\emE{{E}}
\def\emF{{F}}
\def\emG{{G}}
\def\emH{{H}}
\def\emI{{I}}
\def\emJ{{J}}
\def\emK{{K}}
\def\emL{{L}}
\def\emM{{M}}
\def\emN{{N}}
\def\emO{{O}}
\def\emP{{P}}
\def\emQ{{Q}}
\def\emR{{R}}
\def\emS{{S}}
\def\emT{{T}}
\def\emU{{U}}
\def\emV{{V}}
\def\emW{{W}}
\def\emX{{X}}
\def\emY{{Y}}
\def\emZ{{Z}}
\def\emSigma{{\Sigma}}

\newcommand{\etens}[1]{\mathsfit{#1}}
\def\etLambda{{\etens{\Lambda}}}
\def\etA{{\etens{A}}}
\def\etB{{\etens{B}}}
\def\etC{{\etens{C}}}
\def\etD{{\etens{D}}}
\def\etE{{\etens{E}}}
\def\etF{{\etens{F}}}
\def\etG{{\etens{G}}}
\def\etH{{\etens{H}}}
\def\etI{{\etens{I}}}
\def\etJ{{\etens{J}}}
\def\etK{{\etens{K}}}
\def\etL{{\etens{L}}}
\def\etM{{\etens{M}}}
\def\etN{{\etens{N}}}
\def\etO{{\etens{O}}}
\def\etP{{\etens{P}}}
\def\etQ{{\etens{Q}}}
\def\etR{{\etens{R}}}
\def\etS{{\etens{S}}}
\def\etT{{\etens{T}}}
\def\etU{{\etens{U}}}
\def\etV{{\etens{V}}}
\def\etW{{\etens{W}}}
\def\etX{{\etens{X}}}
\def\etY{{\etens{Y}}}
\def\etZ{{\etens{Z}}}

\newcommand{\pdata}{p_{\rm{data}}}
\newcommand{\ptrain}{\hat{p}_{\rm{data}}}
\newcommand{\Ptrain}{\hat{P}_{\rm{data}}}
\newcommand{\pmodel}{p_{\rm{model}}}
\newcommand{\Pmodel}{P_{\rm{model}}}
\newcommand{\ptildemodel}{\tilde{p}_{\rm{model}}}
\newcommand{\pencode}{p_{\rm{encoder}}}
\newcommand{\pdecode}{p_{\rm{decoder}}}
\newcommand{\precons}{p_{\rm{reconstruct}}}

\newcommand{\laplace}{\mathrm{Laplace}} 

\newcommand{\E}{\mathbb{E}}
\newcommand{\Ls}{\mathcal{L}}
\newcommand{\R}{\mathbb{R}}
\newcommand{\emp}{\tilde{p}}
\newcommand{\lr}{\alpha}
\newcommand{\reg}{\lambda}
\newcommand{\rect}{\mathrm{rectifier}}

\newcommand{\sigmoid}{\sigma}
\newcommand{\softplus}{\zeta}
\newcommand{\KL}{\mathrm{KL}}

\newcommand{\Var}{\mathrm{Var}}
\newcommand{\standarderror}{\mathrm{SE}}
\newcommand{\Cov}{\mathrm{Cov}}
\newcommand{\normlzero}{L^0}
\newcommand{\normlone}{L^1}
\newcommand{\normltwo}{L^2}
\newcommand{\normlp}{L^p}
\newcommand{\normmax}{L^\infty}

\newcommand{\parents}{Pa} 

\newcommand{\argmax}{\arg\max}
\newcommand{\argmin}{\arg\min}

\newcommand{\CE}{\mathrm{CE}}
\newcommand{\KLL}{\mathrm{KLloss}}
\newcommand{\Tr}{\mathrm{Tr}}
\newcommand{\TrH}{\mathrm{TrH}}
\newcommand{\diag}{\mathrm{Diag}}
\newcommand{\softmax}{s}
\newcommand{\BCE}{\mathrm{BCE}}
\newcommand{\WKL}{\mathrm{WKL}}

\newcommand{\inp}[1]{\mathcal{I}^{(#1)}_x}
\newcommand{\maxg}[1]{\mathcal{H}^{(#1)}}
\newcommand{\sign}[1]{\mathbb{I}[#1>0]}
\newcommand{\Jprob}{\Phi}
\newcommand{\Jlprob}{\Psi}

\let\ab\allowbreak

\begin{abstract}

Recent research in robust optimization has shown an overfitting-like phenomenon in which models trained against adversarial attacks exhibit higher robustness on the training set compared to the test set. Although previous work provided theoretical explanations for this phenomenon using a robust PAC-Bayesian bound over the adversarial test error, related algorithmic derivations are at best only loosely connected to this bound, which implies that there is still a gap between their empirical success and our understanding of adversarial robustness theory. To close this gap, in this paper we consider
a different form of the robust PAC-Bayesian bound and directly minimize it with respect to the model posterior. The derivation of the optimal solution connects PAC-Bayesian learning to the geometry of the robust loss surface through a Trace of Hessian (TrH) regularizer that measures the surface flatness. In practice, we restrict the TrH regularizer to the top layer only, which results in an analytical solution to the bound whose computational cost does not depend on the network depth. Finally, we evaluate our TrH regularization approach over CIFAR-10/100 and ImageNet using Vision Transformers (ViT) and compare against baseline adversarial robustness algorithms. Experimental results show that TrH regularization leads to improved ViT robustness that either matches or surpasses previous state-of-the-art approaches while at the same time requires less memory and computational cost.

\end{abstract}

\section{Introduction}

\begin{figure}[t]
    \centering
\includegraphics[width=0.51\textwidth]{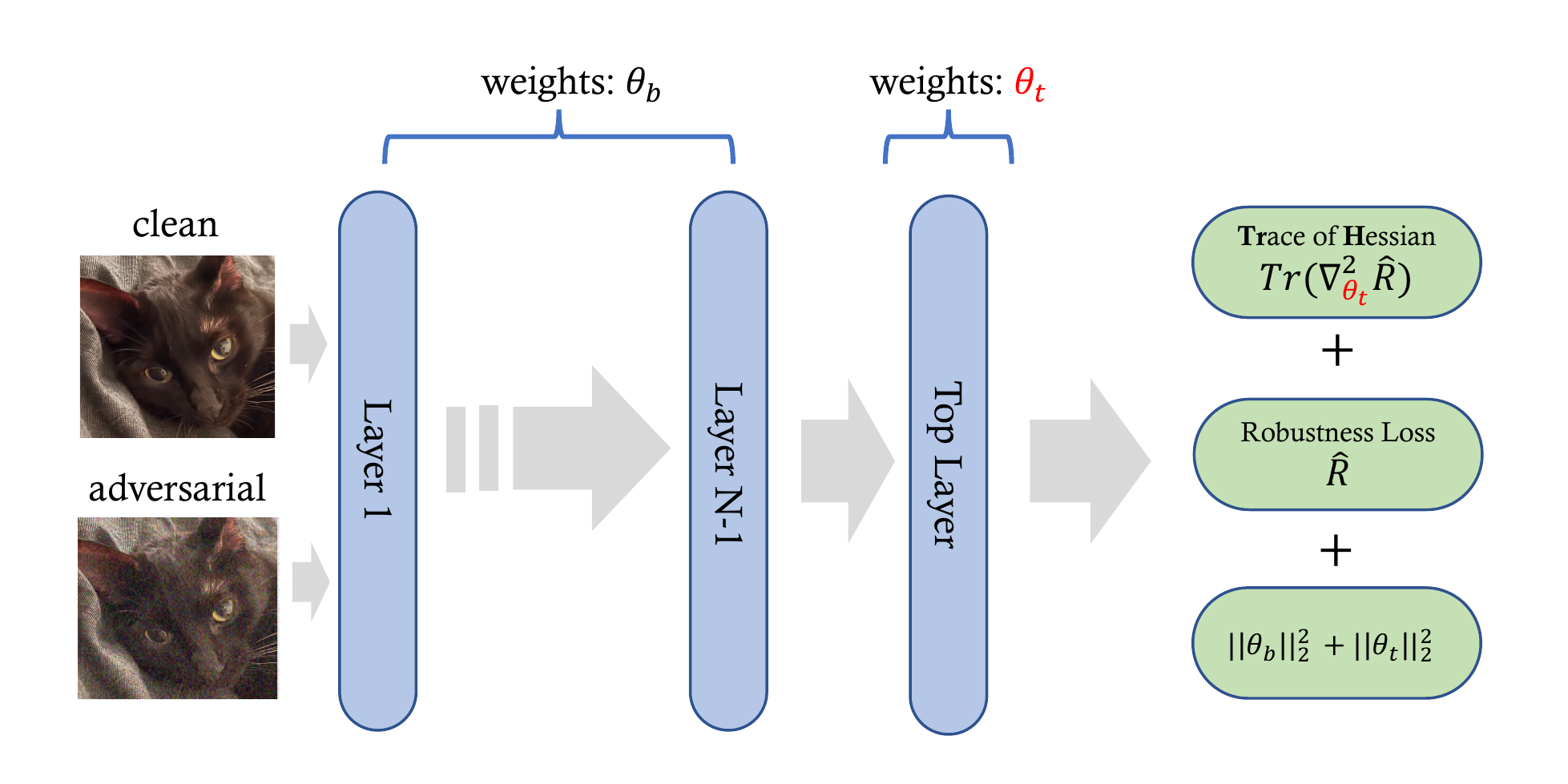}
    \caption{We propose Trace of Hessian (TrH) regularization for training adversarially robust models. In addition to an ordinary robust loss (e.g., TRADES~\cite{zhang2019theoretically}), we regularize the TrH of the loss with respect to the weights of the top layer to encourage flatness. The training objective is the result of direct PAC-Bayesian bound minimization in Theorem~\ref{theorem:trh-bound}.}
    \label{fig:TrH_illustration}
\end{figure}

Despite their success in a wide range of fields and tasks, deep learning models still remain susceptible to manipulating their outputs by even tiny perturbations to the 
input ~\cite{Goodfellow2015ExplainingAH, madry2017towards,carlini2017towards, Kurakin2017AdversarialEI,220580, Wei2019TransferableAA, Carlini2020EvadingDD, Cilloni2022FocusedAA}. 

Several lines of work have focused on developing robust training techniques against such adversarial attacks~\cite{madry2017towards, NEURIPS2018_358f9e7b, zhang2019theoretically, NEURIPS2019_32e0bd14, Wang2020Improving, uncover, leino2021globally, Sehwag2022RobustLM, Pang2022RobustnessAA}. 
Importantly, Rice et al.~\cite{Rice2020OverfittingIA} observe a robust overfitting phenomenon, referred to as the \emph{robust generalization gap}, in which a robustly-trained classifier shows much higher accuracy on adversarial examples from the training set, compared to lower accuracy on the test set.
Indeed, several technical approaches have been developed that could alleviate this overfitting phenomenon, including $\ell_2$ weight regularization, early stopping~\cite{Rice2020OverfittingIA}, label smoothing, data augmentation~\cite{zhang2017mixup, yun2019cutmix}, using synthetic data~\cite{Gowal2021ImprovingRU} and etc. 

According to learning theory, the phenomenon of overfitting can be characterized by a PAC-Bayesian bound~\cite{10.1145/267460.267466, McAllester1999PACBayesianMA, Catoni2004APA, pac-bayesian-linear, PierrePAC} which upper-bounds the expected performance of a random classifier over the underlying data distribution by its performance on a finite set of training points plus some additional terms.

Although several prior works~\cite{wu2020adversarial, Gowal2021ImprovingRU, waIzmailov, Jin_2022_CVPR} have built upon insights from the PAC-Bayesian bound, none attempted to directly minimize the upper bound, 
likely due to the fact that the minimization of their forms of the PAC-Bayesian bound do not have an analytical solution.

In this paper, we rely on a different form of the PAC-Bayesian bound~\cite{pac-bayesian-linear}, which can be readily optimized using a Gibbs distribution~\cite{NIPS2016_84d2004b} with which we derive
a second-order upper bound over the robust test loss.
Interestingly, the resulting bound consists of a regularization term that involves \emph{Trace of Hessian} (TrH)~\cite{pactran} of the network weights, a well-known measure of the loss-surface flatness. 

For practical reasons, we limit TrH regularization to the top layer of the network only because computing a Hessian matrix and its trace for the entire network is too costly. 
We further derive the analytical expression of the top-layer TrH and show both theoretically and empirically that top-layer TrH regularization has a similar effect as regularizing the entire network.
The resulting TrH regularization (illustrated in Figure~\ref{fig:TrH_illustration}) is less expensive and more memory efficient compared to other competitive methods~\cite{wu2020adversarial, Gowal2021ImprovingRU, waIzmailov, Jin_2022_CVPR}.

In summary, our contributions are as follows:
(1) We provide a PAC-Bayesian upper-bound over the robust test loss and show how to directly minimize it (Theorem~\ref{theorem:trh-bound}). To the best of our knowledge, this has not been done by prior work. Our bound includes a TrH term which encourages the model parameters to converge at a flat area of the loss function;
(2) Taking efficiency into consideration, we restrict the TrH regularization to the top layer only (Algorithm~\ref{alg:trh_training}) and show that it is an implicit but empirically effective regularization on the TrH of each internal layer (Theorem~\ref{thm:grad-prop} and Example~\ref{eg:toy_network}); and  
(3) Finally, we conduct experiments with our new TrH regularization and compare the results to several baselines using Vision Transformers~\cite{dosovitskiy2021an}.
On CIFAR-10/100, our method consistently matches or beats the best baseline. 
On ImageNet, we report a significant gain (+2.7\%) in robust accuracy compared to the best baseline and establish a new state-of-the-art result of 48.9\%.

\section{Background}\label{sec:background}
We begin by introducing our notation and basic definitions.
Let $g_\theta(x)$ denote the output logits of a network $\theta$ on an input $x\in \mathcal{X}$. 
We denote the predicted label by $F_\theta(x) = \arg\max g_\theta(x) \in \mathcal{Y}$ and the softmax output of $g_\theta$ by $\softmaxf{g_\theta}$. 
To train the network, we sample a dataset $D^m \defeq \{(x_i, y_i)\}^{m-1}_{i=0}$ containing $m$ i.i.d examples from a distribution $\mathcal{D}$, so that the test and training losses are
\begin{align*}
    & \testloss{\theta} \defeq \E_{(x, y) \sim \mathcal{D}} \;l((x, y), g_\theta),  \\
    \text{and}\;\; & \trainloss{\theta} \defeq \frac{1}{m} \sum^m_{(x, y) \in D^m} l((x, y), g_\theta), 
\end{align*}
respectively and $l$ denotes the loss function.
Ideally, the classification error of $g_\theta$ should be represented by the 0-1 loss $l((x, y), g_\theta) = \mathbb{I}[F_\theta(x) \neq y]$. However, since the 0-1 loss is infeasible to minimize and inferior for generalization, it is common to use surrogate losses such as the Cross-Entropy loss $\CE((x, y), g_\theta) = -\log \{s(g_\theta(x))\}_y$
for training instead, where $\{v\}_j$ denotes the $j$-th element of the vector $v$. 
Finally, we denote the KL-divergence between two distributions $\mathcal{P}, \mathcal{Q}$ as  $\KL(\mathcal{P}||\mathcal{Q})$. For any twice-differentiable function $L$, we denote its Jacobian and Hessian matrix as $\jacobian L$ and $\hessian L$ respectively.


\subsection{Adversarial Robustness}

The goal of \emph{adversarial robustness} is to train a deep network that is robust against $\ell_p$ norm-bounded noises. In this case, the test and training loss functions are replaced by their respective robust counterparts --
 $\robusttestloss{\theta}$ on the test set and  $\robusttrainloss{\theta}$ on the training set, such that
\begin{align}
    &\robusttestloss{\theta} \defeq \E_{(x, y) \sim \mathcal{D}} \max_{||\eps||_p \leq \delta} l((x+\epsilon, y), g_\theta), \nonumber\\
    \text{and }&\robusttrainloss{\theta} \defeq \frac{1}{m} \sum^m_{(x, y) \in D^m} \max_{||\epsilon||_p \leq \delta}l((x+\epsilon, y), g_\theta), \nonumber
\end{align}
where $\delta$ is the maximum noise budget. Various methods have been proposed to improve adversarial robustness. In this paper, we focus on the following two well-known defenses: \emph{adversarial training}~\cite{madry2017towards}, denote AT (Definition~\ref{def:at}), and TRADES~\cite{zhang2019theoretically} (Definition~\ref{def:trades}). 
AT is a classical method in enforcing robustness; while TRADES achieves state-of-the-art robust accuracy on CIFAR-10 and ImageNet~\cite{Gowal2021ImprovingRU}. 

\begin{definition}[Adversarial Training (AT)~\cite{madry2017towards}]\label{def:at}
Given a network with parameter $\theta$, a training set $D^m$, the bound of noise $\delta$ and norm $||\cdot||_p$, AT minimizes the following objective w.r.t $\theta$, 
\begin{align}
    \atloss{\theta} \defeq \frac{1}{m} \sum_{(x, y) \in D^m} \max_{||\epsilon||_p \leq \delta}\CE((x + \epsilon, y), g_\theta). \nonumber 
\end{align}
\end{definition}
which considers the adversarial sample $x+\epsilon$ in a $\delta$-ball around each input, instead of the input itself.

\begin{definition}[TRADES~\cite{zhang2019theoretically}]\label{def:trades}
Given the same assumptions as in Definition \ref{def:at}, TRADES minimizes the following objective w.r.t $\theta$:
\begin{align}
    \tradesloss{\theta} &\defeq \frac{1}{m} \sum_{(x,y)\in D^m} \Big[ \CE((x, y), g_\theta) \nonumber \\ 
    & \;\; + \lambda_t \cdot \max_{||\epsilon||_p \leq \delta}\KLL((x,x+\epsilon), g_\theta) \Big] \nonumber\\
\text{where } \KLL((& x, x+\epsilon), g_\theta) = \KL(s(g_\theta(x)) || s(g_\theta(x+\epsilon)), \nonumber
\end{align}
$\lambda_t$ is a penalty hyper-parameter, which balances the clean accuracy and the robustness.
\end{definition}
Intuitively, TRADES minimizes the cross-entropy loss, while also regularizing for prediction smoothness in an $\delta$-ball around each input.

\subsection{Generalizing Robustness}\label{sec:existing_approaches}
Rice et al.~\cite{Rice2020OverfittingIA} observe that the AT and TRADES methods may suffer from severe overfitting. Namely, the model exhibits higher robustness on the training data compared to the test data. A number of traditional strategies have been applied to alleviate the overfitting problem in robust training, such as $\ell_2$ weight regularization, early stop~\cite{Rice2020OverfittingIA}, label smoothing, data augmentation~\cite{zhang2017mixup, yun2019cutmix} and using synthetic data~\cite{Gowal2021ImprovingRU}. 
Below we describe popular and recent methods that were designed specifically for adversarial training.

\paragraph{Stochastic Weight Averaging (SWA).} 
Izmailov et al.~\cite{waIzmailov} introduces SWA for improved generalization performance in standard training. SWA maintains an exponential running average $\theta_{\texttt{avg}}$ of the model parameters $\theta$ at each training iteration and uses the running average $\theta_{\texttt{avg}}$ for inference. Namely,
\begin{align}
    \theta^{(t+1)} \leftarrow \texttt{SGD}(\theta^{(t)}); \quad \theta_{\texttt{avg}} \leftarrow \alpha \theta_{\texttt{avg}} + (1-\alpha)\theta^{(t+1)} \nonumber 
\end{align}
where $\alpha$ is commonly set to $0.995$.
Recent work has also shown that SWA helps generalization in the robust training~\cite{Gowal2021ImprovingRU} scenario. Note that maintaining the running average requires additional memory resources.

\paragraph{Adversarial Weight Perturbation (AWP).} 
Similar to Sharpness-Aware Minimization~\cite{foret2021sharpnessaware}, Wu et al.~\cite{wu2020adversarial} encourages model robustness by penalizing against the most offending local weight perturbation to the model. 
Specifically, given a robust loss $\rloss{\theta}$,
AWP minimizes the following objective w.r.t $\theta$:
\begin{align}
    \max_{\psi(\xi) \leq \delta_{awp}} \rloss{\theta + \xi} \label{Eq:AWP}
\end{align}
where * denotes either $\texttt{AT}$ or $\texttt{T}$, $\psi$ measures the amount of noise $\xi$ (e.g. $\ell_2$ norm) and $\delta_{awp}$ denotes the noise total budget.
Note that solving the inner maximization problem requires at least one additional gradient ascent step in the parameter space.

\paragraph{Second-Order Statistic (S2O).} 
Jin et al.~\cite{Jin_2022_CVPR} improve the robustness generalization by regularizing the statistical correlations between the weights of the network. Furthermore, they provide a second-order and data-dependent approximation $A_\theta$ of the weight correlation matrix.
During training, S2O minimizes the following objective w.r.t $\theta$:
\begin{align}
    \rloss{\theta} + \alpha ||A_\theta||_F \label{eq:s2o}
\end{align}
where $\alpha>0$ is a hyper-parameter. Computing $A_\theta$ involves computing the per-instance self-Kronecker product of the logit outputs for the clean and adversarial inputs.

\subsection{PAC-Bayesian Theory for Robust Training} \label{sec:background_pac_bayes}
PAC-Bayesian theory~\cite{10.1145/267460.267466, McAllester1999PACBayesianMA, Catoni2004APA, pac-bayesian-linear, PierrePAC,neyshabur2017exploring} provides a foundation for deriving an upper bound of the generalization gap between the training and test error. 
Both AWP and S2O plug the robust loss into the classical bound~\cite{neyshabur2017exploring} to obtain the following form:



\begin{theorem}[Square Root-Form PAC-Bayesian Bound~\cite{wu2020adversarial}]\label{theorem:awp-bound}
Given a prior distribution $\mathcal{P}$ on the weight $\theta$ of a network, any $\tau \in (0, 1]$, a robustness loss $\robusttestloss{\theta}$ for a data distribution $\mathcal{D}$ and its empirical version $\robusttrainloss{\theta}$, for any posterior distribution $\mathcal{Q}$ of $\theta$, the following inequality holds with a probability at least $1-\tau$, 
\begin{align}
    \E_{\theta \in \mathcal{Q}}\robusttestloss{\theta} &\leq  \E_{\theta \in \mathcal{Q}}\robusttrainloss{\theta} \nonumber 
    \\& + 4\sqrt{\frac{1}{m}\KL(\mathcal{Q}||\mathcal{P}) + \log\frac{2m}{\tau}}. \label{Eq:theorem-awp-bound}
\end{align}
\end{theorem}

Generally speaking, the goal of PAC-Bayesian learning is to optimize $\mathcal{Q}$ on the RHS of Eq.~\ref{Eq:theorem-awp-bound} so as to obtain a tight upper bound on the test error (LHS). 
However, directly optimizing Eq.~\ref{Eq:theorem-awp-bound} over $\mathcal{Q}$ is difficult because of the square root term. 
Instead of optimizing the RHS, AWP replaces it with
$\max_{\xi}\robusttrainloss{\theta+\xi}$ plus a constant upper bound on the square root term, which is only loosely connected to the bound.
On the other hand, S2O assumes $\mathcal{P}$ and $\mathcal{Q}$ as non-spherical Gaussian, i.e. $\mathcal{P}=\mathcal{N}(0, \Sigma_p), \mathcal{Q}=\mathcal{N}(\theta, \Sigma_q)$ and show that the square root bound increases as the Frobenius norm and singular value of $\Sigma_q$ increase. To overcome the complexity of the square root term, they approximate it with $\|A_\theta\|_F$ as in Eq.~\ref{eq:s2o}. 

\section{Direct PAC-Bayesian Bound Minimization}\label{sec:method}


As mentioned earlier, neither AWP~\cite{wu2020adversarial} nor S2O~\cite{Jin_2022_CVPR} minimize the PAC-Bayesian bound directly, likely due to the complicating square-root term in Eq.\ref{Eq:theorem-awp-bound}. 
However, Eq.\ref{Eq:theorem-awp-bound} is only one instance out of 
PAC-Bayesian bounds defined in Germain et al.~\cite{pac-bayesian-linear}. 
According to their framework, there is also a \emph{linear-form} of the bound related to Bayesian inference which can be analytically minimized with the aid of a Gibbs distribution~\cite{germain2016pac,rothfuss2021pacoh,pactran}. 
We leverage this linear-form of the PAC-Bayesian bound, and adapt it for bounding the robustness gap:

\begin{theorem}[Linear-Form PAC-Bayesian Bound~\cite{germain2016pac}]\label{theorem:linear-bound}
Under the same assumptions as in Theorem \ref{theorem:awp-bound}, for any $\beta > 0$, with a probability at least $1-\tau$, 
\begin{align}
\E_{\theta \in \mathcal{Q}}\robusttestloss{\theta} \leq & \;\E_{\theta \in \mathcal{Q}}\robusttrainloss{\theta} \nonumber\\ 
&+ \frac{1}{\beta}\KL(\mathcal{Q}||\mathcal{P})+C(\tau, \beta, m), \label{eq:pac-bayesian-robustness} 
\end{align}
where $C(\tau, \beta, m) $ is a function independent of $\mathcal{Q}$.
\end{theorem}
Here $\beta$ is a hyper-parameter that balances the three terms on the RHS. A common choice is $\beta \propto m$, i.e. the size of the dataset~\cite{germain2016pac, NEURIPS2021_f6b6d2a1}.
Recently, \cite{pactran} proposed to set $\mathcal{P}$ and $\mathcal{Q}$ to univariate Gaussians and used a second-order approximation to estimate the PAC-Bayesian bound. 
Applying a similar technique, we introduce Theorem~\ref{theorem:trh-bound} (see Appendix.~\ref{appendix:proof} for the proof) which minimizes the RHS of Eq.~\ref{eq:pac-bayesian-robustness}.

\begin{theorem}[Minimization of PAC-Bayesian Bound]\label{theorem:trh-bound}
If $\mathcal{P} = \mathcal{N}(\mathbf{0}, \sigma^2_0)$, and $\mathcal{Q}$ is also a product of univariate Gaussian distributions, then the minimum of Eq.~\ref{eq:pac-bayesian-robustness} w.r.t $\mathcal{Q}$ can be bounded by
\begin{align}
    &\min_{\mathcal{Q}} \E_{\theta \in \mathcal{Q}}\robusttestloss{\theta} \nonumber\\
    \leq & \min_{\mathcal{Q}}\{ \E_{\theta \in \mathcal{Q}}\robusttrainloss{\theta} + \frac{1}{\beta}\KL(\mathcal{Q}||\mathcal{P})\} +C(\tau, \beta, m) \nonumber \\
    = &\min_{\theta} \{ \robusttrainloss{\theta}  + \frac{||\theta||^2_2}{2\beta \sigma^2_0}  +\frac{\sigma^2_0}{2}\Tr(\triangledown^2_{\theta}\robusttrainloss{\theta}) \} \nonumber\\
    & \quad +C(\tau, \beta, m) + O(\sigma^4_0).
    \label{eq:minimal_of_the_bound}
\end{align}
\end{theorem}


Assuming that $\sigma_0^2 = 1/d$ is the variance of the initial weight matrices where $d$ is the input feature dimension, then $O(\sigma^4_0) = O(d^{-2})$.
Theorem~\ref{theorem:trh-bound} considerably simplifies the optimization problem from one over the space of probability density functions $\mathcal{Q}$ to one over model weights $\theta$. 
Moreover, the resulting bound (RHS of Eq.~\ref{eq:minimal_of_the_bound}) contains the Trace of Hessian (TrH) term 
$\Tr(\triangledown^2_{\theta}\robusttrainloss{\theta})$ of the robust loss, which sums the loss surface curvatures in all directions.
For a convex loss, the Hessian is positive semi-definite (PSD) in which case trace minimization leads $\theta$ to a flatter region of the loss surface.
Although in general, our Hessian is not PSD, empirically we find that its largest eigenvalue has a much larger magnitude than the least ones~\cite{alain2018negative} 
and that trace minimization correlates with minimizing the standard deviation of the eigenvalues (see Figure \ref{fig:eigen-spectrum} in the appendix), so that the overall curvature is effectively decreases.


The RHS of Eq.~\ref{eq:minimal_of_the_bound} in Theorem~\ref{theorem:trh-bound} provides a training objective function that bounds the optimal test error up to a constant. However, evaluating and minimizing the bound in Eq.~\ref{eq:minimal_of_the_bound} requires computing the Trace of Hessian (TrH) term.
Unfortunately, computing the Hessian directly by applying auto-differentiation twice is infeasible for a large deep network. Another solution is to apply Hutchinson's method~\cite{10.1145/1944345.1944349} to randomly approximate the TrH,
however, the variance of the resulting estimator is too high and it ends up being too noisy for training in practice.

Instead, we propose to restrict TrH regularization to the top layer of the deep network only. Although this restriction brings approximation error to the bound, it has two benefits.
First, the TrH on the top layer has simple analytical expressions for  (Section \ref{sec:regularization-linear}). Second, we show that the top-layer TrH regularization effectively regularizes the TrH of the entire network as well (Section \ref{sec:regularization-deep}). 





\subsection{Top-Layer TrH regularizers for AT \& TRADES}\label{sec:regularization-linear}

The network parameter $\theta$ can be decomposed into $\{\theta_t, \theta_b\}$ where $\theta_t$ denotes the weights of the top layer and $\theta_b$ denotes the weights of the remaining (lower) network. In this case, the logits can be expressed as $g_\theta(x) =\theta^\top_t f_{\theta_b}(x)$, where $f_{\theta_b}(x)$ is the penultimate layer feature. Furthermore, we define $h(x, \theta) = \softmax({g}_\theta(x))-\softmax({g}_\theta(x))^2$. Now, we present the analytical forms of the top-layer TrH w.r.t the AT and the TRADES in Proposition \ref{prop:at_trh} and \ref{prop:trades_trh}. 


\begin{proposition}[TrH for AT]\label{prop:at_trh} Given a training dataset $D^m$ and the adversarial input example $x'$ for each example $x$, the top-layer TrH of the AT loss (Definition \ref{def:at}) is equal to
\begin{align}
\Tr(\triangledown^2_{\theta_t} \atloss{\theta}) = \frac{1}{m} \sum_{(x, y) \in D^m} \TrH_{\texttt{AT}}(x';\theta), \nonumber\\
\text{where }\;\; \TrH_{\texttt{AT}}(x';\theta)= ||f_{\theta_b}(x')||^2_2 \cdot \mathbf{1}^\top h(x', \theta) \nonumber.
\end{align} 
\end{proposition}

\begin{proposition}[TrH for TRADES]\label{prop:trades_trh} Under the same assumption in Proposition~\ref{prop:at_trh}, the top-layer TrH of the TRADES loss (Definition \ref{def:trades}) is equal to
\begin{align}
    &\Tr(\triangledown^2_{\theta_t} \tradesloss{\theta}) = \frac{1}{m} \sum_{(x, y) \in D^m} \TrH_{\texttt{T}}(x, x'; \lambda_t, \theta), \nonumber\\
&\text{where,}\;\; \TrH_{\texttt{T}}(x, x'; \lambda_t, \theta) = ||f_{\theta_b}(x)||^2_2 \cdot \mathbf{1}^\top h(x, \theta) \nonumber\\
&\qquad\;\;\;\;\;\;\;\;\;\; +\lambda_t ||f_{\theta_b}(x')||^2_2 \cdot \mathbf{1}^\top h(x', \theta)  \label{eq:traded_trh}. 
\end{align}
    
\end{proposition}
To obtain Eq.~\ref{eq:traded_trh}, we stopped the gradient on the $\KLL$ in Definition~\ref{def:trades} with respect to the clean logits $g_\theta(x)$, because we find it is more stable for training, otherwise, there would be an additional regularization term $G(x, x', \theta)$ in the $\TrH_{\texttt{T}}$ (see more details in the Appendix~\ref{appendix:proof}).


With Proposition~\ref{prop:at_trh} and~\ref{prop:trades_trh}, we now present the complete training objective in Algorithm~\ref{alg:trh_training}. Notice that, to simplify hyper-parameter notations, we re-parameterize $\gamma \defeq 1/2\beta\sigma^2_0$ and $\lambda \defeq \sigma^2_0/2$ in the algorithm. 




\begin{algorithm2e*}[t]
\SetAlgoLined
\caption{TrH Regularization for Training Adversarial Robust Network}\label{alg:trh_training}
\small
\KwHp{The $\ell_2$-norm penalty for weights $\gamma$, the TRADES penalty $\lambda_t$ and the TrH penalty $\lambda$.}
\KwIn{A batch $B$ of examples, 
a \texttt{loss\_type} $= $'$\texttt{AT}$' or '$\texttt{TRADES}$', the noise budget $\delta$, and a network $g_\theta=\theta^{\top}_t f_{\theta_b}(x)$.}
\KwOut{The training objective at the current iteration.}
$R \gets 0;$\\
\ForEach{$(x, y) \in B$}{
$x' \gets \texttt{PGD\_InnerLoop}(x, y, \delta$, \texttt{loss\_type} )\\
$z' \gets f_{\theta_b}(x'), \quad h' \gets s(\theta^{\top}_t z') - s(\theta^{\top}_t z')^2;$ \tcp*[r]{$s$ is the softmax function. This step computes the penultimate feature $z'$ and the softmax gradient $h'$ on the adversarial input $x'$.}
\uIf{\texttt{loss\_type} == `$\texttt{AT}$'}{
$R \gets R + \CE((x', y), g_\theta) + \lambda( ||z'||^2_2 \cdot \mathbf{1}^\top h')$;}
\ElseIf{\texttt{loss\_type} == `$\texttt{TRADES}$'}{
$z \gets f_{\theta_b}(x), \quad h \gets s(\theta^{\top}_t z) - s(\theta^{\top}_t z)^2;$ \tcp*[r]{TRADES needs the $z$ and $h$ for the clean input $x$.}
$R \gets R + \CE((x, y), g_\theta) + \lambda_t \KL(s(\theta^{\top}_t z)||s(\theta^{\top}_t z')) + \lambda( ||z||^2_2 \cdot \mathbf{1}^\top h+\lambda_t ||z'||^2_2 \cdot \mathbf{1}^\top h');$ }}
\KwReturn{$\frac{R}{|B|}+\gamma||\theta||^2_2$}
\end{algorithm2e*}

\subsection{The effect of Top-Layer TrH Regularization}\label{sec:regularization-deep}


Although the TrH regularization is restricted to the top layer only, the gradient back-propagates to the entire network through the penultimate layer features. This motivates the following research question:

\textit{What effect does top-layer TrH regularization have on the TrH of the entire network?} 


To answer this question, we take a two-step approach. First, we experiment with a small 3-layer network on the synthetic Two-Moon dataset (see Appendix~\ref{appendix:efficiency-example} for a visualization of the dataset), provided that computing the full network Hessian is inexpensive. Subsequently, we provide a theorem that can be used to bound the Trace of Hessian of the full network with that of the top layer only.

\begin{figure}[t]
    \centering
\includegraphics[width=0.48\textwidth]{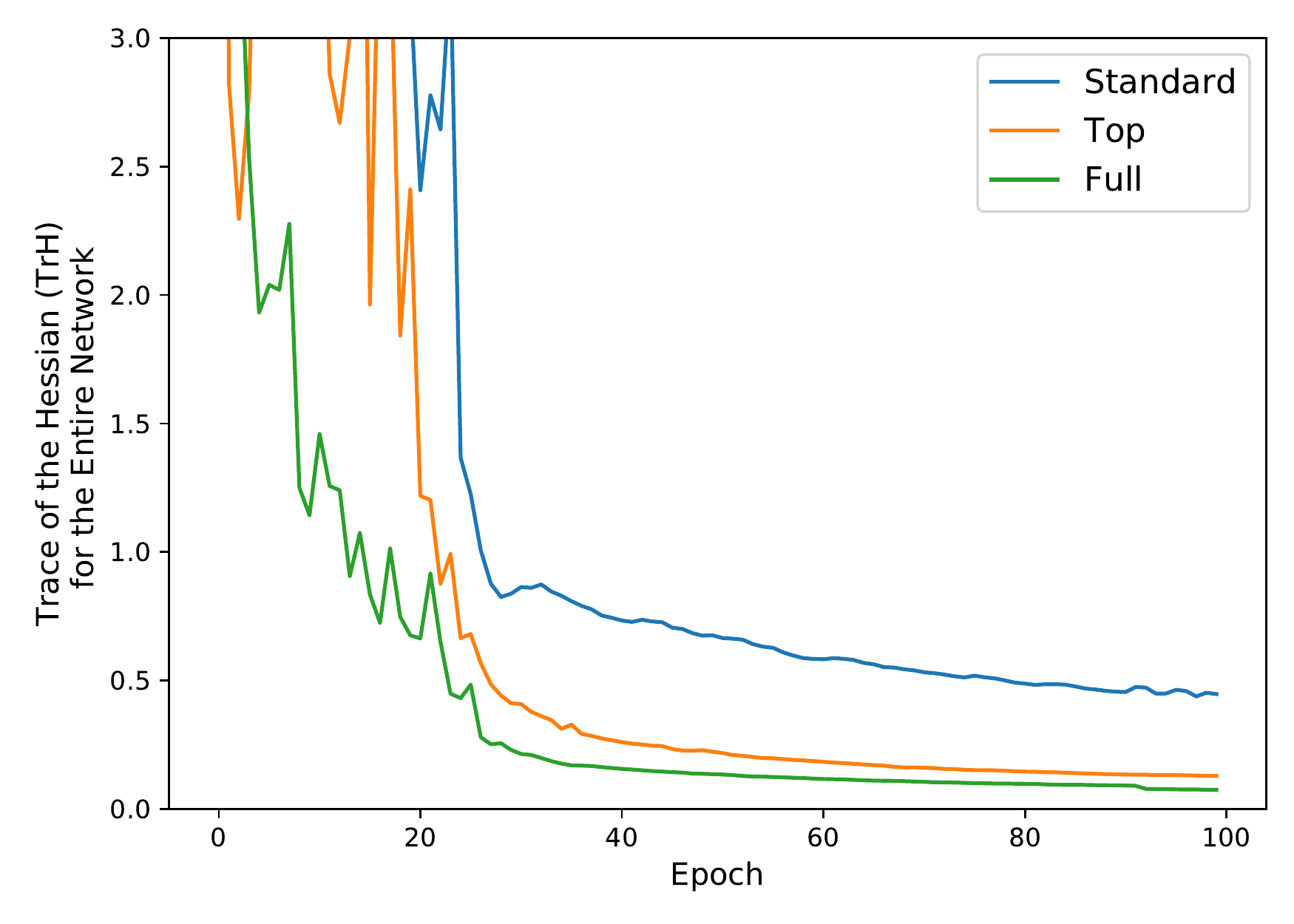}
    \caption{Trace of the Hessian (TrH) of the entire network (y-axis) over training epochs (x-axis) with (1) no regularization (Standard, in blue); (2) top layer TrH regularization (Top, in orange); and (3) regularization the TrH of all layers (Full, in green). See Example~\ref{eg:toy_network} for the experimental setup.}
    \label{fig:toy_example}
\end{figure}

\begin{example}[Two Moons]\label{eg:toy_network} We use $500$ training examples, where $x \in \mathbb{R}^2$ and $y \in \{0, 1\}$  from the Two Moon dataset. We train a 3-layer fully-connected network with Dense(100)-ReLU-Dense(100)-ReLU-Dense(2) with the AT loss under three settings:
\begin{itemize}
    \item Standard: no regularization.
    \item Top: Regularizing the TrH of the top layer only.
    \item Full: Regularizing the TrH of the entire network. 
\end{itemize}
Throughout the training process (x-axis), we evaluate the TrH of the \emph{entire} network (y-axis) with a numerical approach (twice differentiation) and plot the results in Figure~\ref{fig:toy_example}.
\end{example}

Figure~\ref{fig:toy_example} empirically shows that regularizing the TrH of the top layer (orange) indirectly reduces the TrH of the entire network and is nearly as effective as regularizing the entire network (green) after the 60-th epoch. 

Next, we provide a theoretical justification for the impact of top-layer TrH regularization on the layers below. Intuitively, our Theorem~\ref{thm:grad-prop}, establishes an inductive relation between the TrH of the consecutive layers in a feedforward neural network with ReLU activation and CE loss, a common building block in AT and TRADES training.

\begin{theorem}[Inductive Relation in TrH]\label{thm:grad-prop}
Suppose $g_\theta$ is a feed-forward network with ReLU activation. For the $i$-th layer, let $W^{(i)}$ be its weight and $\inp{i} $ be its input evaluated at $x$. Thus,  $\TrH^{\CE}_x(W^{(i-1)})$, i.e. TrH evaluated at $x$ using a CE loss w.r.t $W^{(i-1)}$, is equal to
\begin{align}
    \TrH^{\CE}_x(W^{(i-1)}) = ||\{\inp{i-1}\}||^2_2  \sum_{k, d\in P^{(i)}} \{\maxg{i}\}_{k, d} \nonumber
\end{align}
\begin{align}
\text{where }\{\maxg{i}\}_{k, d_i} &\defeq \left[ \frac{\partial \{g_\theta(x)\}_k}{\partial \{\inp{i}\}_{d_i}}  \right]^2 \cdot \{h(x, \theta)\}_k \nonumber \\ P^{(i)} &\defeq \{d_i | \{\inp{i}\}_{d_i} > 0\}  \nonumber
\end{align}
and the following inequality holds for any $\maxg{i}, \maxg{i+1}$:
\begin{align}
   \max_{k, d_i} \{\maxg{i}\}_{k, d_i} \leq \max_{k, d_{i+1}} \{\maxg{i+1}\}_{k, d_{i+1}} \cdot ||W^{(i)}||^2_1.  \label{eq:operator-norm-of-weights}
\end{align} 
\end{theorem}


Put simply, the max-norm of the Hessian of each layer forms an upper bound for the max-norm of the layer below it (times a scalar). Therefore, regularizing the TrH of the top layer only implicitly regularizes the Hessian of the layers below it. This effect explains the phenomenon observed in Example~\ref{eg:toy_network}. 

Furthermore, the operator norm $||W^{(i)}||^2_1$ of weights in Eq.~\ref{eq:operator-norm-of-weights} makes an interesting connection between TrH minimization and robust training. Roth et al.~\cite{NEURIPS2020_ab731488} show that robust training is an implicit way of regularizing the weight operator norm. Moreover, directly minimizing the operator norm is a commonly used technique in training certifiable robust networks~\cite{leino2021globally, huang2021training}, a stronger version of the robustness guarantee targeted in this paper. Thus, the combination of robustness training and top-layer regularization results in an implicit regularization on TrH for internal layers. 

\section{Evaluation}\label{sec:evaluation}
In this section, we evaluate TrH regularization (Algorithm~\ref{alg:trh_training}) 
over three image classification benchmarks including CIFAR-10/100 and ImageNet and compare
against the robust training baselines described in Section~\ref{sec:existing_approaches}. 

\begin{table*}[!t]
\centering

\begin{tabular}{lcccc|cccc}
\\
\toprule
\toprule
$\ell_\infty (\delta= 8/255)$  & \multicolumn{4}{c|}{ViT-L16}                            & \multicolumn{4}{c}{Hybrid-L16}       \\ 
\footnotesize{SE}$=\se{0.5}$\% &
  \multicolumn{2}{c}{CIFAR-10} &
  \multicolumn{2}{c|}{CIFAR-100} &
  \multicolumn{2}{c}{CIFAR-10} &
  \multicolumn{2}{c}{CIFAR-100} \\ 
Defense &
  \texttt{Clean}(\%) &
  \texttt{AA}(\%) &
  \texttt{Clean} (\%)&
  \multicolumn{1}{c|}{\texttt{AA}(\%)} &
  \texttt{Clean}(\%) &
  \texttt{AA}(\%) &
  \texttt{Clean} (\%)&
  \texttt{AA}(\%) \\ \midrule
AT(\base)     &87.4& \multicolumn{1}{c|}{60.8} &64.3& \multicolumn{1}{c|}{31.7} & 88.0& \multicolumn{1}{c|}{\textbf{61.7}} & 64.2 & \underline{31.8}\\
AT(\WA)    &88.0& \multicolumn{1}{c|}{\underline{61.7}} &62.5& \multicolumn{1}{c|}{31.7} & 86.5& \multicolumn{1}{c|}{57.2} & 63.6 & 30.8 \\
AT(\STO)   &87.1& \multicolumn{1}{c|}{60.2} &64.9& \multicolumn{1}{c|}{31.7} & 88.7& \multicolumn{1}{c|}{\underline{61.6}} & 64.3 & \underline{31.8}\\
AT(\AWP)     &88.5& \multicolumn{1}{c|}{\textbf{62.4}} &63.3& \multicolumn{1}{c|}{\underline{32.6}} &88.5 & \multicolumn{1}{c|}{\underline{61.5}} & 63.9 & \textbf{32.5}\\
AT(\texttt{TrH})    &87.0& \multicolumn{1}{c|}{\underline{61.7}} &62.5& \multicolumn{1}{c|}{\textbf{32.8}} & 88.0& \multicolumn{1}{c|}{\underline{61.0}} & 63.8 & \underline{32.4}\\ \midrule
TRADES(\base) &85.2& \multicolumn{1}{c|}{60.3} &62.0& \multicolumn{1}{c|}{\underline{31.4}} &85.9 & \multicolumn{1}{c|}{\underline{60.8}} & 61.3 & 30.6\\
TRADES(\WA)  &85.9& \multicolumn{1}{c|}{\underline{60.7}} &61.5& \multicolumn{1}{c|}{\underline{32.0}} &84.3 & \multicolumn{1}{c|}{59.4} & 62.3 & 29.9 \\
TRADES(\STO)  &87.4& \multicolumn{1}{c|}{\textbf{61.6}} &62.4& \multicolumn{1}{c|}{\underline{31.6}} & 87.2 & \multicolumn{1}{c|}{\underline{61.5}} & 65.0 & 31.9\\
TRADES(\AWP)  &85.3& \multicolumn{1}{c|}{\underline{60.8}} &63.0& \multicolumn{1}{c|}{\textbf{32.2}} & 84.5& \multicolumn{1}{c|}{59.8} & 61.4 & 32.1\\
TRADES(\texttt{TrH}) &86.4& \multicolumn{1}{c|}{\underline{61.4}} &62.0& \multicolumn{1}{c|}{\underline{32.1}} &87.4 & \multicolumn{1}{c|}{\textbf{61.7}} & 66.4 & \textbf{34.1} \\ \bottomrule
\end{tabular}

\centering
\begin{tabular}{lcccc|cccc}
\\
\toprule
\toprule
ImageNet  & \multicolumn{4}{c|}{ViT-B16}                            & \multicolumn{4}{c}{ViT-L16}       \\ 
 \footnotesize{SE}$=\se{0.2}$\% &
  \multicolumn{2}{c}{$\ell_\infty (\delta=4/255)$} &
  \multicolumn{2}{c|}{$\ell_2 (\delta=3.0)$} &
  \multicolumn{2}{c}{$\ell_\infty (\delta=4/255)$} &
  \multicolumn{2}{c}{$\ell_2 (\delta=3.0) $} \\ 
Defense &
  \texttt{Clean}(\%) &
  \texttt{AA}(\%) &
  \texttt{Clean} (\%)&
  \multicolumn{1}{c|}{\texttt{AA}(\%)} &
  \texttt{Clean}(\%) &
  \texttt{AA}(\%) &
  \texttt{Clean} (\%)&
  \texttt{AA}(\%) \\ \midrule
AT(\base)     & 72.2 & \multicolumn{1}{c|}{39.0} & 71.0 & \multicolumn{1}{c|}{39.3} & 75.4 & \multicolumn{1}{c|}{44.1} &  73.9& 43.5 \\
AT(\WA)       & 72.3 & \multicolumn{1}{c|}{40.0} & 71.5 & \multicolumn{1}{c|}{39.8} & 75.2& \multicolumn{1}{c|}{44.3} & 74.3 & 43.8 \\
AT(\STO)   & 72.0 & \multicolumn{1}{c|}{39.0} & 71.4 & \multicolumn{1}{c|}{39.3} & 75.4 & \multicolumn{1}{c|}{46.2} & 76.3 & 46.4 \\
AT(\AWP)      &  71.3& \multicolumn{1}{c|}{39.5} & 70.5 & \multicolumn{1}{c|}{39.3} & 74.5& \multicolumn{1}{c|}{44.0} &73.8  & 44.0 \\
AT(\texttt{TrH})      &  71.3& \multicolumn{1}{c|}{\textbf{42.2}} & 71.6 & \multicolumn{1}{c|}{\textbf{42.4}} & 75.8 & \multicolumn{1}{c|}{\textbf{48.8}} & 74.2& \textbf{47.0} \\ \midrule
TRADES(\base) & 69.2 & \multicolumn{1}{c|}{39.3} & 67.8 & \multicolumn{1}{c|}{39.7} & 77.8 & \multicolumn{1}{c|}{40.9} & 77.1 & 41.5 \\
TRADES(\WA)   & 69.5 & \multicolumn{1}{c|}{39.7} & 68.1 & \multicolumn{1}{c|}{40.0} & 72.7 & \multicolumn{1}{c|}{44.5} & 71.1 & 44.2 \\
TRADES(\STO) & 70.6 &  \multicolumn{1}{c|}{39.3} & 69.8 & 39.7 & 73.7& \multicolumn{1}{c|}{43.6} & 72.5 &  44.0  \\
TRADES(\AWP)  & 65.8 & \multicolumn{1}{c|}{38.6} & 64.5 & \multicolumn{1}{c|}{38.3} &  68.5& \multicolumn{1}{c|}{43.2} & 67.2 &  42.7\\
TRADES(\texttt{TrH}) &  65.3& \multicolumn{1}{c|}{\textbf{41.9}} & 66.4 & \multicolumn{1}{c|}{\textbf{41.6}} &
70.1  & \multicolumn{1}{c|}{\textbf{48.9}} & 68.9  &  \textbf{47.3}\\ 
\bottomrule
\end{tabular}

\caption{\texttt{Clean}: \% of Top-1 correct predictions. \texttt{AA}: \% of Top-1 correct predictions under AutoAttack. A max Standard Error (SE)~\cite{stark2005sticigui} = $\sqrt{0.5*(1-0.5)/m}$ ($m$ as the number of test examples) is computed for each dataset. The best results appear in bold. Underlined results are those that fall within the SE range of the result and are regarded roughly equal to the best result.}
\label{tab:main_results}

\end{table*}
\subsection{Experiment Setup}\label{sec:evaluation-setup}
\paragraph{Datasets and Threat Model.} For CIFAR-10/100, we choose the standard $\epsilon=0.031 (\approx 8/255)$ of the $\ell_\infty$ ball to bound the adversarial noise. 
In addition, Gowal et al.~\cite{Gowal2021ImprovingRU} has reported that using synthetic images from a DDPM~\cite{ho2020denoising} generator can improve the robustness accuracy for the original test set. Thus, we also add 1M DDPM images to CIFAR-10/100 (these images are publicly availabe~\cite{deepmind}).
For ImageNet, we choose $\epsilon=3.0$ for the $\ell_2$ and $\epsilon=0.0157 (\approx 4/255)$ for the $\ell_\infty$ ball, respectively. We do not use extra data for ImageNet. 
The chosen $\epsilon$s follow the common choices in the literature~\cite{Gowal2021ImprovingRU, wu2020adversarial, madry2017towards, Jin_2022_CVPR}.  We report the test accuracy evaluated with benign images and adversarial images generated by AutoAttack~\cite{croce2020reliable} (AutoPGD+AutoDLR), which is widely used in the literature~\cite{Gowal2021ImprovingRU, wu2020adversarial, Jin_2022_CVPR}. 

\paragraph{Network Architectures.} We use Vision Transformer (ViT)~\cite{dosovitskiy2021an} through all experiments because of its success on multiple tasks over existing architectures. Specifically, we report the results of ViT-L16, Hybrid-L16 for CIFAR-10/100 and ViT-B16, ViT-L16 for ImageNet (additional results and architecture details can be found in Appendix~\ref{appendix:architecture}). 
We initialize the weights from a pre-trained checkpoint on ImageNet21K~\cite{google-research}. Notice that the resolution of the CIFAR images ($32\times 32$) are too small to correctly align with the pre-trained kernel of the first layer. To address this issue, we down-sample the kernel of the input layer following the receipt proposed by Mahmood et al.~\cite{Mahmood_2021_ICCV}. 

\paragraph{Methods.} We train the models using AT and TRADES losses, together with the following defenses: (1) base: no regularization; (2) AWP~\cite{wu2020adversarial}; (3) SWA~\cite{waIzmailov}; (4) S2O~\cite{Jin_2022_CVPR}; and (5) our Algorithm~\ref{alg:trh_training} (TrH). 
We are interested in the baselines defenses because of their connections to the PAC-Bayesian bound or the loss flatness.
During training, the model was partitioned over four Google Cloud TPUv4 chips~\cite{Jouppi2017IndatacenterPA} for the base and TrH methods, and eight chips for AWP, S2O, and SWA as they consume more HBM memory.

\paragraph{Hyperparameter Selection.} Training robust ViTs can be challenging due to a large number of hyper-parameters. We use two steps for hyperparameter selection. 
For the choice of common hyper-parameters, e.g. batch size, patch size, training iterations, and etc. (a full list is included in Appendix~\ref{appendix:pre-tuning}), we first do a large grid search and select values that produce the best results on TRADES(\base). This step is referred to as pre-tuning and is done per dataset.


After the first step, where the common hyper-parameters are locked, we tune and report the best results after trying different sets of method-specific hyper-parameters, e.g., the norm bound of weight noises in AWP and $\lambda$ for TrH. Appendix~\ref{appendix:fine-tuning} provides a full list of hyper-parameters we use to produce our main results in Table~\ref{tab:main_results}. 

\subsection{Main Results}\label{sec:evaluation-result}

In Table~\ref{tab:main_results}, we report the robust test accuracy using different types of defenses. The top results are highlighted in bold font. To measure the significance of an improvement, we calculate the maximum standard error (SE)~\cite{stark2005sticigui} = $\sqrt{0.5*(1-0.5)/m}$ (where  $m$ denotes the number of test examples) for each dataset. 
Thus, the accuracy of a certain model is regarded a \emph{silver} result $a_{silver}$, if its SE interval overlaps with the that of the top result $a_{top}$. Namely, $a_{top} - SE \leq a_{silver} + SE$.



\paragraph{CIFAR-10/100.}
In CIFAR-10/100, a significant improvement is at least 1\% according to SE. Therefore, based on Table~\ref{tab:main_results}, the performance differences among the methods are relatively small. Nevertheless, TrH attains either the top result (in 3 instances) or the silver result (in 5 instances) across all eight instances. In comparison, AWP is the second best with 3 top and 3 silver; S2O has 1 top and 4 silver; SWA has 3 silver; and the base has 1 top and 3 silver. 
Notably, Hybird-L16+TRADES(\texttt{TrH}) achieves the highest robust accuracy (34.1\%) on CIFAR-100 which beats all other baselines by at least 2\%. 

\paragraph{ImageNet.} On ImageNet, a significant improvement is at least 0.4\% according to SE. We observe that TrH regularization outperforms the other baselines across the board. Using AT(\texttt{TrH}), we set a new \emph{state-of-the-art} robust accuracy of 48.8\%, at $\ell_\infty (4/255)$ and 47.0\% at $\ell_2 (3.0)$ with a ViT-L16 model.
Specifically, in the case of $\ell_\infty$, 48.8\% is an improvement of 4.7\% compared to the basic setup AT(\base). 
In TRADES(\texttt{TrH}), although the robust accuracy is even slightly higher than AT, the clean accuracy is lower. This is due to the choice of $\lambda_t=6$, which strongly favors robustness over clean accuracy. As we reduce $\lambda_t$ to 1, we obtain a better balance of (clean, AA) accuracy of TRADES(\texttt{TrH}) at (78.7, 46.7) for $l_\infty$ and (77.1, 45.2) for $l_2$.

\paragraph{Summary} The results in Table~\ref{tab:main_results} demonstrate that training with TrH regularization either matches or outperforms the robust accuracy of the existing methods.
In fact, the gains of TrH regularization are wider on the larger ImageNet dataset compared to on CIFAR-10/100, where several methods show equivalent performance (per Standard Error analysis).
This suggests that future work in robust adversarial training should move beyond the CIFAR benchmarks to larger-scale tasks.

\subsection{Sensitivity and Efficiency}

\begin{figure}
    \centering
\includegraphics[width=0.44\textwidth]{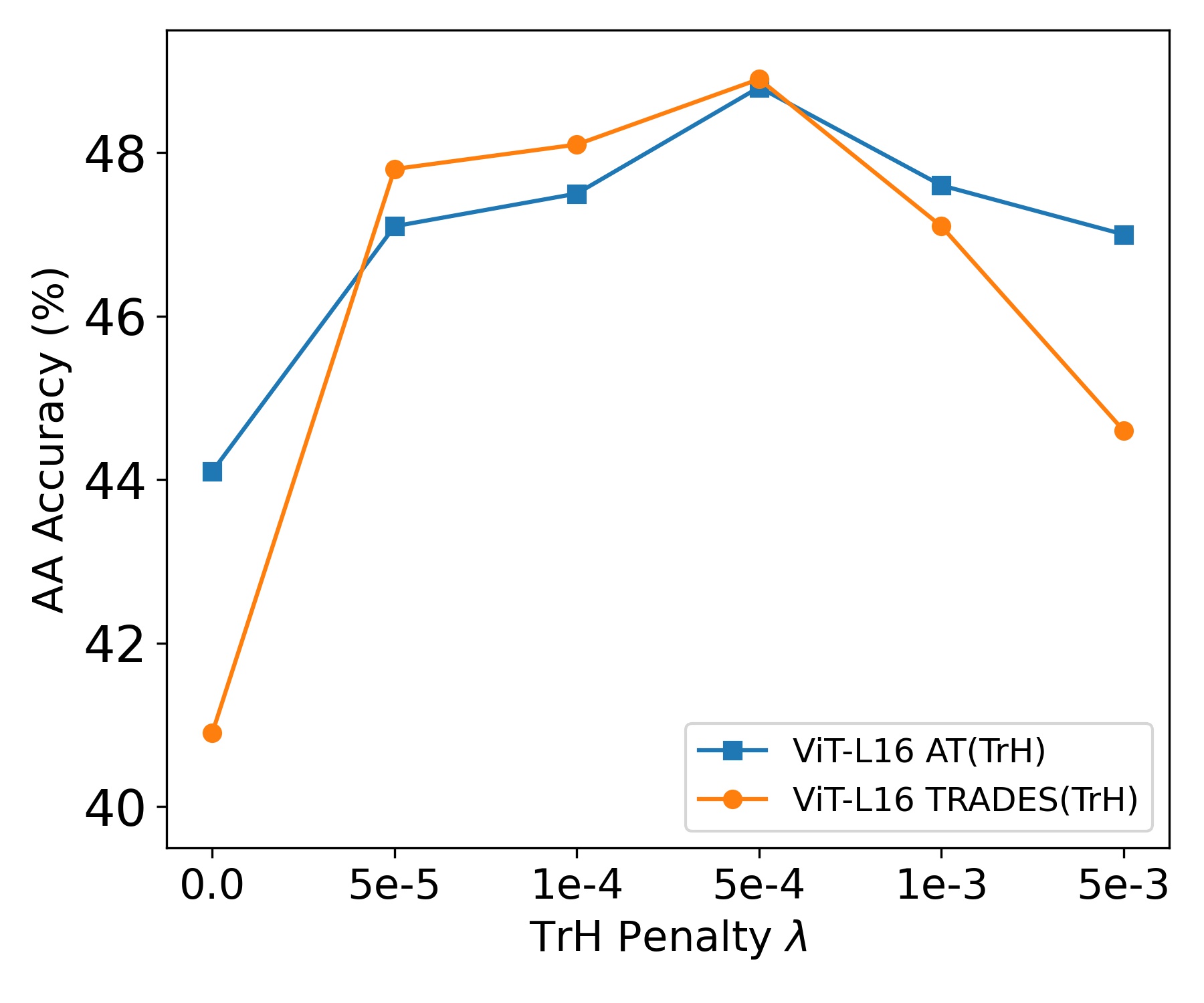}
    \caption{A plot of the Autoattack accuracy (\texttt{AA} \%) against $\lambda$ (TrH Penalty coefficient) when training on ImageNet with TrH Regularization in the $\ell_\infty$ case.}
    \label{fig:sensitivity}
\end{figure}

\paragraph{Sensitivity to $\lambda$.} To study the sensitivity of $\lambda$ to the training in TrH regularization, we train a ViT-L16 model over the ImageNet dataset and vary the choice of $\lambda$. 
Figure~\ref{fig:sensitivity}.
plots the \texttt{AA} accuracy (y-axis) against the different choices of $\lambda$ (x-axis) with all other setups fixed. 
The plot shows that $\lambda=5\times10^{-4}$ attains the highest \texttt{AA} accuracy for both AT and TRADES losses with little degradation in the range of $[10^{-4}, 10^{-3}]$. 

\paragraph{Training Efficiency.} To compare the computational efficiency of the various methods, we report the peak memory usage and runtime performance (i.e., images processed per second; the higher the better) in Table~\ref{fig:runtime-report}. As the cloud TPU chips are dynamically allocated and preemptable, we instead measure these costs on two local NIVIDA RTX chips with 24G memory per chip. Because by default JAX preallocates all GPUs, we set \texttt{XLA\_PYTHON\_CLIENT\_ALLOCATOR=platform} before measuring the memory allocation and run-time. In Table~\ref{fig:runtime-report}, we show that adding TrH regularization brings almost no additional memory usage and runtime slowdown, compared to training without any regularization. 

\newcommand{\fakeimage}{{\fboxsep=-\fboxrule\fbox{\rule{0pt}{3cm}\hspace{4cm}}}}



\begin{table}[t]
\centering
\begin{tabular}{lcc}
\toprule
\toprule

Defense &
  Mem.(MB/chip) &
  Img/sec/chip \\ \midrule
AT(\base)     &$5.0 \times10^3$&5.5\\
AT(\texttt{TrH}) (ours)     & $5.0\times10^3$ & 5.2 \\
AT(\WA)     & $5.6\times10^3$ & 4.4\\
AT(\STO)     & $8.0  \times10^3$& 5.2\\
AT(\AWP)     &  $6.8\times10^3$& 3.7\\
 \bottomrule
\end{tabular}
\caption{Peak memory usage and run-time reports measured when training CIFAR-10 using a ViT-L16 with a batch size of 32. Training is paralleled on two NVIDIA RTX chips.  Lower memory usage and higher img/sec/chip are more efficient.}
\label{fig:runtime-report}
\end{table}

\section{Related Work and Discussion}\label{sec:related_work}
In this section, we situate our contributions in the broader context of current literature, such as related work on PAC-Bayesian theory, flatness regularization, as well as other types of robust adversarial losses.

\paragraph{PAC-Bayesian Theory and Flatness.} 
Beyond the PAC-Bayesian inspired approaches for robust training that were already mentioned in the paper (such as Wu et al.~\cite{wu2020adversarial}), a different line of work focuses on adapting PAC-Bayesian bounds for an ensemble of models, to improve either adversarial robustness~\cite{NEURIPS2021_78e8dffe} or out-of-distribution (OOD) robustness data~\cite{Zecchin2022RobustPT}.
Several other works have focused on methods for finding flat local minima of the loss surface in order to improve (standard) model performance\cite{foret2021sharpnessaware, Jia2020,pactran}.
Also outside the scope of adversarial robustness, Ju et al.~\cite{pmlr-v162-ju22a} find that trace of Hessian regularization leads to robustness against noise in the \emph{labels}. 
Their Hessian regularization is based on randomized estimation and is done for all layers, whereas we effectively apply TrH regularization on the top layer only and with Theorem~\ref{thm:grad-prop} as theoretical justification. 

Perhaps the closest approach to ours is the PACTran-Gaussian metric~\cite{pactran}, which uses a 2nd-order approximation of the PAC-Bayesian bound as the metric for evaluating the transferability of pretrained checkpoints. Similarly to our Eq.~\ref{eq:minimal_of_the_bound}, the PACTran metric is composed of two terms: an $\ell_2$ regularized empirical risk (RER) and a flatness regularizer (FR). Our work was inspired by the PACTran metric, but also makes several improvements. Firstly, in PACTran a single posterior variance $\sigma_q^2$ was shared for all parameters. On the contrary, we used different $\sigma_q^2$'s for different parameters. Secondly, in PACTran the optimization of the posterior mean $\theta_q$ is over the RER term only, while our optimization is over both RER and FR. These two changes potentially make our bound tighter than the one in the PACTran paper. In addition, we also study and explicitly include the high-order terms in $O(\sigma_0^4)$, which was neglected in PACTran.




\paragraph{Adversarial Surrogate Loss.} 

In this paper, we focused only on the AT and TRADES approaches for robust adversarial optimization as they achieve reasonable results on the datasets of choice.
In fact, recent work~\cite{Gowal2021ImprovingRU} on achieving state-of-the-art robustness on CIFAR-10 also exclusively focuses on AT or TRADES.
Beyond AT and TRADES, the works in ~\cite{Wang2020Improving, Ding2020MMA, Pang2022RobustnessAA} explore other types of adversarial surrogate losses. Our method can be easily transferred to these losses and we include additional examples in Appendix~\ref{appendix:other-loss}.




\section{Conclusion}

In summary, we alleviate overfitting in adversarial training with PAC-Bayesian theory.
Using a linear-form PAC-Bayesian bound overlooked by the prior work, we derive a 2nd-order estimation that includes the Trace of Hessian (TrH) of the model.
We then focus on the TrH of the top layer only, showing both empirically and theoretically (for ReLU networks) that its minimization effectively leads to the minimization of the TrH of the entire network, and thereby providing a sound flatness regularization technique for adversarial training which adds only little computational overhead.
Experimental results show that our TrH regularization method is consistently among the top performers on CIFAR-10/100 benchmark as well as achieves a significant improvement in robust test accuracy on ImageNet compared to the other baselines.

{\small
\bibliographystyle{ieee_fullname}
\bibliography{egbib}
}
\clearpage
\appendix
\onecolumn
\begin{appendices}
\section{Theorems and Proofs}\label{appendix:proof}

\subsection{Proof of Theorem~\ref{theorem:trh-bound}}
\paragraph{Theorem~\ref{theorem:trh-bound}}
If $\mathcal{P} = \mathcal{N}(\mathbf{0}, \sigma^2_0)$, and $\mathcal{Q}$ is also a product of univariate Gaussian distributions, then the minimum of Eq.~\ref{eq:pac-bayesian-robustness} w.r.t $\mathcal{Q}$ can be bounded by
\begin{align*}
    &\min_{\mathcal{Q}} \E_{\theta \in \mathcal{Q}}\robusttestloss{\theta} \nonumber\\
    \leq & \min_{\mathcal{Q}}\{ \E_{\theta \in \mathcal{Q}}\robusttrainloss{\theta} + \frac{1}{\beta}\KL(\mathcal{Q}||\mathcal{P})\} +C(\tau, \beta, m) \nonumber \\
    = &\min_{\theta} \{ \robusttrainloss{\theta}  + \frac{||\theta||^2_2}{2\beta \sigma^2_0}  +\frac{\sigma^2_0}{2}\Tr(\triangledown^2_{\theta}\robusttrainloss{\theta}) \} +C(\tau, \beta, m) + O(\sigma^4_0).
\end{align*}

\textit{Proof.} Before we begin our proof, we emphasize that the posterior $\mathcal{Q}$ that minimizes the test loss may not be the same posterior $\mathcal{Q}$ that minimizes the training loss. Namely, define
\begin{align*}
    \mathcal{Q}_{test} &\defeq  \arg\min_{\mathcal{Q}} \E_{\theta \in \mathcal{Q}}\robusttestloss{\theta}\\
    \mathcal{Q}_{train} &\defeq  \arg\min_{\mathcal{Q}}\{ \E_{\theta \in \mathcal{Q}}\robusttrainloss{\theta} + \frac{1}{\beta}\KL(\mathcal{Q}||\mathcal{P})\},
\end{align*} 
then $\mathcal{Q}_{test}$ may not be equal to $\mathcal{Q}_{train}$. However, the following inequality clearly holds based on their definitions,
\begin{align*}
    \E_{\theta \in \mathcal{Q}_{test}}\robusttestloss{\theta} \leq \E_{\theta \in \mathcal{Q}_{train}}\robusttestloss{\theta} \leq \E_{\theta \in \mathcal{Q}_{train}}\robusttrainloss{\theta} + \frac{1}{\beta}\KL(\mathcal{Q}||\mathcal{P}) +C(\tau, \beta, m).
\end{align*}
Generally speaking, it is impossible to directly find $\mathcal{Q}_{test}$ without knowing the true data distribution $\mathcal{D}$.
Thus, Theorem~\ref{theorem:trh-bound} attempts to derive $\mathcal{Q}_{train}$ to minimize the RHS of Eq.~\ref{eq:pac-bayesian-robustness}. When it is clear which optimal posterior, i.e. $\mathcal{Q}_{train}$ instead of $\mathcal{Q}_{test}$, we can derive in the theorem, we omit the subscription and directly write $\mathcal{Q}$.

\paragraph{Proof Overview.} Based on our assumptions, we write $\mathcal{Q} = \mathcal{N}(\theta, \Sigma)$ where $\Sigma$ is the (diagonal) covariance matrix. 
The minimization of the bound w.r.t $\mathcal{Q}$ is to minimize the bound w.r.t $\Sigma$ and $\theta$. Our proof is therefore three-fold: (1) firstly, we write the expression of $\KL(Q||P)$ using $\theta$ and $\Sigma$; (2) secondly, we use a second-order Taylor expansion to decompose $\E_{\theta \in \mathcal{Q}}\robusttrainloss{\theta}$  into terms as functions of $\Sigma$ and $\theta$; and (3) finally, we minimize the bound w.r.t $\Sigma$ for two cases: all weights have the same or different variances. Namely, the diagonal of $\Sigma$ has the same or different elements. At the end of the minimization w.r.t $\Sigma$, we arrive at the bound that requires to minimize $\theta$ to actually minimize the RHS. Our proof follows below.

\paragraph{Extra Notations.} Throughout the proof we will use $N$ for the total number of weights, i.e. the cardinality of $\theta$. When indexing a particular weight, we write $\theta_n$. Let the diagonal of $\Sigma$ be $\sigma^2 = [\sigma^2_1, \sigma^2_2, \cdots, \sigma^2_N ]^\top$, i.e. the variance of each weight. We use $\diag(a)$ to denote the diagonal matrix where the the diagonal is the elements from vector $a$.

\paragraph{Step I: the KL term $\KL(\mathcal{Q}||\mathcal{P})$.} 
The expression of the KL term $\KL(\mathcal{Q}||\mathcal{P})$ when $\mathcal{P} = \mathcal{N}(\mathbf{0}, \sigma^2_0I)$ is as follows, 
\begin{align}
    \KL(\mathcal{Q}||\mathcal{P}) &= \frac{1}{2}\left[ \log \frac{\sigma^2_0}{\det(\Sigma)} - N + \frac{||\theta||^2_2}{\sigma^2_0} + \frac{\Tr(\Sigma)}{\sigma^2_0}\right] \nonumber \\
    &=\frac{1}{2}\left[\sum_n \log \frac{\sigma^2_0}{\sigma^2_n} - N + \frac{||\theta||^2_2}{\sigma^2_0} + \frac{\sum_n \sigma^2_n} {\sigma^2_0}\right] (\text{when $\Sigma$ is a diagonal matrix}). \label{eq:kl-multivariate-gaussian}
\end{align}

\paragraph{Step II: Second-order Taylor's Expansion for $\E_{\theta \in \mathcal{Q}}\robusttrainloss{\theta}$.} We expand $\E_{\theta \in \mathcal{Q}}\robusttrainloss{\theta}$ at the point $\theta$ and use the re-parameterization trick,
\begin{align}
   \E_{\theta \in \mathcal{Q}}\robusttrainloss{\theta} &= \robusttrainloss{\theta} + \underbrace{\mathbb{E}_{\epsilon \sim \mathcal{N}(0, \Sigma)} [\epsilon^\top \triangledown \robusttrainlosss]}_{=0 \text{ because $\epsilon$ has zero mean} } + \frac{1}{2}  \mathbb{E}_{\epsilon \sim \mathcal{N}(0, \Sigma)} [\epsilon^\top \triangledown^2 \robusttrainlosss\epsilon ] \nonumber \\
   &+ \underbrace{\frac{1}{6}  \mathbb{E}_{\epsilon \sim \mathcal{N}(0, \Sigma)} [\sum_{i, j, k} \frac{\partial^3 \robusttrainlosss}{\partial \theta_i \theta_j \theta_k}\epsilon_i\epsilon_j\epsilon_k ]}_{=0 \text{ because the mean and the skewness of a standard Gaussian are 0}} + O(\sigma^4) \nonumber\\
   &=\robusttrainloss{\theta}+  \frac{1}{2}  \mathbb{E}_{\epsilon \sim \mathcal{N}(0, I)} [(\sigma \circ \epsilon)^\top \triangledown^2 \robusttrainlosss(\sigma \circ \epsilon) ] + O(\sigma^4) \label{eq:second-order-expansion-with-circ}
\end{align}
where $\circ$ is the element-wise product, a.k.a the Hadamard product. Notice the connection between a vector product and Hadamard product:
\begin{align}
    \text{for any vectors }a, b, \quad a \circ b = \diag(a)b.\label{eq:hadamard-dot-relation}
\end{align}
Using Eq.~\ref{eq:hadamard-dot-relation}, we simplify Eq.~\ref{eq:second-order-expansion-with-circ} as follows,
\begin{align}
    \E_{\theta \in \mathcal{Q}}\robusttrainloss{\theta} &=\robusttrainloss{\theta}  + \frac{1}{2}  \mathbb{E}_{\epsilon \sim \mathcal{N}(0, I)} [(\diag(\sigma) \epsilon)^\top \triangledown^2 \robusttrainlosss(\diag(\sigma) \epsilon) ] + O(\sigma^4) \nonumber \\
    &=\robusttrainloss{\theta} + \frac{1}{2}  \Tr (\Sigma \circ \triangledown^2 \robusttrainlosss ) + O(\sigma^4). \label{eq:second-order-expansion-with-tr}
\end{align}

\paragraph{Step III: Bound Minimization.} Using Eq.~\ref{eq:kl-multivariate-gaussian} and~\ref{eq:second-order-expansion-with-tr}, we re-write the bound in Theorem~\ref{theorem:linear-bound} as follows
\begin{align}
    E_{\theta \in \mathcal{Q}}\robusttrainloss{\theta} &\leq \E_{\theta \in \mathcal{Q}}\robusttrainloss{\theta} + \frac{1}{\beta}\KL(\mathcal{Q}||\mathcal{P})+C(\tau, \beta, m) \nonumber \\
    &=\robusttrainloss{\theta} + \frac{1}{2}  \Tr (\Sigma \circ \triangledown^2 \robusttrainlosss ) + \frac{1}{2\beta}\left[\sum_n \log \frac{\sigma^2_0}{\sigma^2_n} - N + \frac{||\theta||^2_2}{\sigma^2_0} + \frac{\sum_n \sigma^2_n}{\sigma^2_0}\right] + O(\sigma^4) + C(\tau, \beta, m). \label{eq:expansion-of-linear-bound}
\end{align}
There are two sets of parameters in Eq.~\ref{eq:expansion-of-linear-bound} for bound minimization: $\Sigma$ (i.e. the $\sigma^2$) and $\theta$.
To minimize w.r.t $\Sigma$, strictly speaking it requires minimizing all relevant terms and the higher-order terms in $O(\sigma^4)$, which is clearly infeasible. 
Instead, we apply the following approximation by neglecting the higher-order $O(\sigma^4)$ and solve
\begin{align*}
    \sigma^{*2} = \arg\min_{\sigma^2}\Big\{ \robusttrainloss{\theta}+ \frac{1}{2}  \Tr (\Sigma \circ \triangledown^2 \robusttrainlosss ) + \frac{1}{2\beta}\left[\sum_n \log \frac{\sigma^2_0}{\sigma^2_n} - N + \frac{||\theta||^2_2}{\sigma^2_0} + \frac{\sum_n \sigma^2_n}{\sigma^2_0}\right]\Big\}.
\end{align*}

In the following, we discuss two cases of $\sigma_n^2$.
\paragraph{Case I (Spherical Gaussian): $\forall n, \sigma^2_{n} = \sigma^2_{q}$.} In this case, the solution to the problem above is to take the derivative w.r.t. $\sigma^2_{q}$ and set it to zero. That is, for the optimal $\sigma^{*2}_q$, 
\begin{align*}
\Tr(\robusttrainloss{\theta}) + \frac{K}{\beta}(\frac{1}{\sigma^2_0} - \frac{1}{\sigma^{*2}_q}) = 0 \implies \frac{\sigma^2_0}{\sigma^{*2}_q} = 1+\frac{\sigma^2_0\beta}{K}\Tr(\robusttrainloss{\theta}).
\end{align*}
Substituting each $\sigma^2_n$ with $\sigma^{*2}_q$ in Eq.~\ref{eq:expansion-of-linear-bound} gives the solution to the following problem
\begin{align*}
    \min_{\sigma^2} \left[ \E_{\theta \in \mathcal{Q}}\robusttrainloss{\theta} + \frac{1}{\beta}\KL(\mathcal{Q}||\mathcal{P})\right] = \robusttrainloss{\theta} + \frac{K}{2\beta} \log(1+\frac{\sigma^2_0\beta}{K}\Tr(\robusttrainloss{\theta})) + \frac{||\theta||^2_2}{2\beta\sigma^2_0}  + O(\sigma^{*4}_q).
\end{align*}
With the Taylor expansion for $\log(1+x) = x + O(x^2)$, 
\begin{align*}
    \frac{K}{2\beta} \log(1+\frac{\sigma^2_0\beta}{K}\Tr(\robusttrainloss{\theta})= \frac{\sigma^2_0}{2\beta}\Tr(\robusttrainloss{\theta}) + O(\sigma^4_0) + O(\sigma^{*4}_q). 
\end{align*}
We can actually merge $O(\sigma^4_0) + O(\sigma^{*4}_q)$ because 
\begin{align*}
    \frac{\sigma^2_0}{\sigma^{2*}_q} = 1+\frac{\sigma^2_0\beta}{K}\Tr(\robusttrainloss{\theta}) \implies \sigma^2_0 \geq \sigma^{*2}_q
\end{align*}
so $O(\sigma^4_0) + O(\sigma^{*4}_q) = O(\sigma^4_0)$. Lastly, to minimize the bound, we optimize it w.r.t $\theta$, which leads us to land on:
\begin{align*}
    \min_{\theta} \{ \robusttrainloss{\theta}  + \frac{||\theta||^2_2}{2\beta \sigma^2_0}  +\frac{\sigma^2_0}{2}\Tr(\triangledown^2_{\theta}\robusttrainloss{\theta}) \} +C(\tau, \beta, m) + O(\sigma^4_0).
\end{align*}

\paragraph{Case II (Diagonal): $\exists n_1, n_2, \sigma^2_{n_1} \neq \sigma^2_{n_2}$.} In this case, we take the derivative w.r.t each $\sigma^2_n$ and set it to zero. That is, for each optimal $\sigma^{2*}_n$,
\begin{align*}
\frac{\partial \robusttrainlosss^2}{\partial \theta^2_n} + \frac{1}{\beta}(\frac{1}{\sigma^2_0} - \frac{1}{\sigma^{2*}_n}) = 0 \implies \frac{\sigma^2_0}{\sigma^{2*}_n} = 1+\sigma^2_0\beta\frac{\partial^2 \robusttrainlosss}{\partial \theta^2_n}.
\end{align*}
Substituting $\sigma^2_n$ with $\sigma^{2*}_n$ in Eq.~\ref{eq:expansion-of-linear-bound} gives the solution to the following problem
\begin{align*}
    \min_{\sigma^2_n} \left[ \E_{\theta \in \mathcal{Q}}\robusttrainloss{\theta} + \frac{1}{\beta}\KL(\mathcal{Q}||\mathcal{P})\right] = \robusttrainloss{\theta}  + \frac{1}{2\beta}\sum_n \log(1+\sigma^2_0\beta\frac{\partial \robusttrainlosss^2}{\partial \theta^2_n}) + \frac{||\theta||^2_2}{2\beta\sigma^2_0}  + O(\sigma^{*4}).
\end{align*}
Similar to Case I, we take Taylor's Expansion of $\log(1+x)$ so 
\begin{align*}
    \frac{1}{2\beta}\sum_n \log(1+\sigma^2_0\beta\frac{\partial \robusttrainlosss^2}{\partial \theta^2_n}) =\frac{\sigma^2_0}{2}\Tr(\robusttrainloss{\theta}) + O(\sigma^4_0).
\end{align*}
Lastly, using $O(\sigma^4_0) + O(\sigma^{*4}) = O(\sigma^4_0)$, we land on the same objective as derived in Case I:
\begin{align*}
    \min_{\theta} \{ \robusttrainloss{\theta}  + \frac{||\theta||^2_2}{2\beta \sigma^2_0}  +\frac{\sigma^2_0}{2}\Tr(\triangledown^2_{\theta}\robusttrainloss{\theta}) \} +C(\tau, \beta, m) + O(\sigma^4_0).
\end{align*}

\subsection{Proof of Theorem~\ref{thm:grad-prop}}
\noindent\textbf{Theorem~\ref{thm:grad-prop}} Suppose that $g_\theta$ is a feed-forward network with ReLU activation. For the $i$-th layer, let $W^{(i)}$ be its weight and $\inp{i} \in \mathbb{R}^{D^{(i)}} $ be its input evaluated at $x$. Thus,  $\TrH^{\CE}_x(W^{(i-1)})$, i.e. TrH evaluated at $x$ using a CE loss w.r.t $W^{(i-1)}$, is as follows
\begin{align}
    \TrH^{\CE}_x(W^{(i-1)}) = ||\{\inp{i-1}\}||^2_2  \sum_{k, d\in P^{(i)}} \{\maxg{i}\}_{k, d} \nonumber
\end{align}
\begin{align}
\text{where }\{\maxg{i}\}_{k, d_i} &\defeq \left[ \frac{\partial \{g_\theta(x)\}_k}{\partial \{\inp{i}\}_{d_i}}  \right]^2 \cdot h_k(x, \theta) \nonumber \\ P^{(i)} &\defeq \{d_i | \{\inp{i}\}_{d_i} > 0\}  \nonumber
\end{align}
and the following inequality holds for any $\maxg{i}, \maxg{i+1}$:
\begin{align}
   \max_{k, d_i} \{\maxg{i}\}_{k, d_i} \leq \max_{k, d_{i+1}} \maxg{i+1} \cdot ||W^{(i)}||^2_1.  \nonumber
\end{align}

\textit{Proof.}
We use $d_{i-1}, d_{i}$ and $d_{i+1}$ to index the element of $\inp{i-1}$, $\inp{i}$ and $\inp{i+1}$. By the definition of trace, we need to compute the second-order derivative of the CE loss w.r.t to each entry in $W^{(i)}$. That is, we write
\begin{align}
    \TrH^{\CE}_x(W^{(i-1)}) = \sum_{d_{i-1}, d_i}\frac{\partial^2 }{\partial W^{(i-1)2}_{d_{i-1}, d_i}}(CE((x, y), g_\theta)) \label{eq:trh-ce-wi}.
\end{align}
First, we derive the first-order derivative. The following steps are based on the chain rule:
\begin{align*}
     \frac{\partial }{\partial W^{(i-1)}_{d_{i-1}, d_i}} (CE((x, y), g_\theta)) &= \sum_k \frac{\partial }{\partial \{g_\theta(x)\}_k} (CE((x, y), g_\theta)) \cdot \frac{\{g_\theta(x)\}_k}{\partial W^{(i-1)}_{d_{i-1}, d_i}}\\
     &= \sum_k (\{s(g_\theta(x)\}_k - y_k) \cdot \frac{\{g_\theta(x)\}_k}{\partial W^{(i-1)}_{d_{i-1}, d_i}}.
\end{align*}
The second-order derivative is therefore equal to
\begin{align*}
    \frac{\partial^2 }{\partial W^{(i-1)2}_{d_{i-1}, d_i}}(CE((x, y), g_\theta)) &= \sum_k \Big(\frac{\partial}{\partial W^{(i-1)}_{d_{i-1}, d_i}} \left[\{s(g_\theta(x)\}_k - y_k)\right] \cdot \frac{\{g_\theta(x)\}_k}{\partial W^{(i-1)}_{d_{i-1}, d_i}} + h_k(x, \theta) \cdot \frac{\partial}{\partial W^{(i-1)}_{d_{i-1}, d_i}} \left[ \frac{\{g_\theta(x)\}_k}{\partial W^{(i-1)}_{d_{i-1}, d_i}}\right]\Big)\\
    &= \sum_k \Big(h_k(x, \theta) \cdot \left[\frac{\{g_\theta(x)\}_k}{\partial W^{(i-1)}_{d_{i-1}, d_i}}\right]^2 + h_k(x, \theta) \cdot \frac{\partial}{\partial W^{(i-1)}_{d_{i-1}, d_i}} \left[ \frac{\{g_\theta(x)\}_k}{\partial W^{(i-1)}_{d_{i-1}, d_i}}\right]\Big).
\end{align*}
Notice that the first-order derivative of ReLU is either 1 or 0, so we have plenty of identity functions in $\frac{\{g_\theta(x)\}_k}{\partial W^{(i-1)}_{d_{i-1}, d_i}}$. Rigorously, ReLU does not have a second-order derivative. In practice, one will return 0 when querying the second-order derivative of ReLU. Thus, the following holds in practice.
\begin{align}
    \frac{\partial}{\partial W^{(i-1)}_{d_{i-1}, d_i}} \left[ \frac{\partial \{g_\theta(x)\}_k}{\partial W^{(i-1)}_{d_{i-1}, d_i}}\right] = 0. \label{eq:relu-grad-trick}
\end{align}
Eq.~\ref{eq:relu-grad-trick} simplifies the expression of the second-order derivative w.r.t $W^{(i-1)}_{d_{i-1}, d_i}$ so we have the following clean expression:
\begin{align}
   \frac{\partial^2 }{\partial W^{(i-1)2}_{d_{i-1}, d_i}}(CE((x, y), g_\theta)) &= \sum_{k} \left[\frac{\partial \{g_\theta(x)\}_k}{\partial W^{(i-1)}_{d_{i-1}, d_i}}\right]^2 \cdot h_k(x, \theta)\nonumber\\
   &= \sum_{k} \left[ \frac{\partial \{g_\theta(x)\}_k}{\partial \{{\inp{i}}\}_{d_i}} \cdot \frac{\partial \{\inp{i}\}_{d_i}}{\partial W^{(i-1)}_{d_{i-1}, d_i}} \right]^2 \cdot h_k(x, \theta) \nonumber\\
   & = \sum_{k} \left[ \frac{\partial \{g_\theta(x)\}_k}{\partial \{\inp{i}\}_{d_i}} \cdot \sign{\{\inp{i}\}_{d_i}} \cdot \{\inp{i-1}\}_{d_{i-1}}  \right]^2 \cdot h_k(x, \theta)\nonumber\\
   &= \sign{\{\inp{i}\}_{d_i}} \cdot \{\inp{i-1}\}^2_{d_{i-1}} \cdot \sum_{k} \left[ \frac{\partial \{g_\theta(x)\}_k}{\partial \{\inp{i}\}_{d_i}}  \right]^2 \cdot h_k(x, \theta).
   \label{eq:second-order-derivative-wi}
\end{align}
By defining $\{\maxg{i}\}_{k,d_i} \defeq \left[ \frac{\partial \{g_\theta(x)\}_k}{\partial \{\inp{i}\}_{d_i}}  \right]^2 \cdot h_k(x, \theta)$, we further simplify Eq.~\ref{eq:second-order-derivative-wi} as follows
\begin{align}
    \frac{\partial^2 }{\partial W^{(i-1)2}_{d_{i-1}, d_i}}(CE((x, y), g_\theta)) &= \sign{\{\inp{i}\}_{d_i}}\cdot \{\inp{i-1}\}^2_{d_{i-1}} \cdot \sum_{k} \{\maxg{i}\}_{k,d_i}.
    \label{eq:second-order-derivative-wi-with-H}
\end{align}
We plug Eq.~\ref{eq:second-order-derivative-wi-with-H} back to Eq.~\ref{eq:trh-ce-wi} and arrive

\begin{align*}
    \TrH^{\CE}_x(W^{(i-1)}) &= \sum_{d_{i-1}, d_i}  \sign{\{\inp{i}\}_{d_i}} \cdot \{\inp{i}\}^2_{d_{i-1}} \cdot \sum_{k} \{\maxg{i}\}_{k,d_i}\\
    &= \Big( \sum_{d_{i-1}}  \{\inp{i-1}\}^2_{d_{i-1}}\Big) \cdot \sum_{k, d_i}\sign{\{\inp{i}\}_{d_i}} \cdot \{\maxg{i}\}_{k,d_i}\\
    &= ||\{\inp{i-1}\}||^2_2 \cdot \sum_{k, d_i}\sign{\{\inp{i}\}_{d_i}} \cdot \{\maxg{i}\}_{k,d_i}.
\end{align*}
By defining $P^{(i)}$ as a set of indices of positive neurons in $\inp{i}$, i.e, $\forall d_2 \in P^{(i)},\{\inp{i}\}_{d_2} > 0$, we simplify the equation above as follows
\begin{align*}
\TrH^{\CE}_x(W^{(i-1)}) = ||\{\inp{i}\}||^2_2 \cdot \sum_{k, d_i\in P^{(i)}} \{\maxg{i}\}_{k,d_i}.
\end{align*}
Furthermore, by the definition of $\maxg{i}$, we notice that
\begin{align} 
    \{\maxg{i}\}_{k,d_i} &=\left[ \frac{\partial \{g_\theta(x)\}_k}{\partial \{\inp{i}\}_{d_i}}  \right]^2 \cdot h_k(x, \theta)\nonumber\\
    &= \Biggl\{\sum_{d_{i+1}} \left[\frac{\partial \{g_\theta(x)\}_k}{\partial \{\inp{i+1}\}_{d_{i+1}}} \cdot \frac{\partial \{\inp{i+1}\}_{d_{i+1}}}{\partial \{\inp{i}\}_{d_i} } \right] \Biggr\}^2\cdot  h_k(x, \theta) \label{eq: before-devision-by-nonzero}\\
    &= \Biggl\{\sum_{d_{i+1}} \left[ \Big(\frac{\{\maxg{i+1}\}_{k,d_{i+1}}}{h_k(x, \theta)}\Big)^{\frac{1}{2}} \cdot \frac{\partial \{\inp{i+1}\}_{d_{i+1}}}{\partial \{\inp{i}\}_{d_i} } \right] \Biggr\}^2\cdot  h_k(x, \theta). \label{eq: after-devision-by-nonzero}
\end{align}
Notice that the transition from Eq.~\ref{eq: before-devision-by-nonzero} to Eq.~\ref{eq: after-devision-by-nonzero} is because $\forall k, h_k(x, \theta) > 0$. Thus, we find
\begin{align}
     \{\maxg{i}\}_{k,d_i} &= \Biggl\{\sum_{d_{i+1}} \left[\{\maxg{i+1}\}^{\frac{1}{2}}_{k,d_{i+1}} \cdot \frac{\partial \{\inp{i+1}\}_{d_{i+1}}}{\partial \{\inp{i}\}_{d_i} } \right] \Biggr\}^2.\label{eq:H_k,d-with-out-max}
\end{align}
For each layer $i$, we denote $\maxg{i}_{max} = \max_{k, d_{i}}  \{\maxg{i}\}_{k, d_{i}}$. Since $\forall i, d_{i},  \{\maxg{i}\}_{k, d_{i}} > 0$, we can take the square root for $\maxg{i}_{max}$ such that 
\begin{align*}
    \forall k, d_{i}, (\maxg{i}_{max})^{\frac{1}{2}} \geq (\{\maxg{i}\}_{k, d_{i}})^{\frac{1}{2}}. 
\end{align*}
Thus, we can bound $ \{\maxg{i}\}_{k,d_i}$ as follows:
\begin{align*}
    \{\maxg{i}\}_{k,d_i} &\leq \Biggl\{\sum_{d_{i+1}} \left[(\maxg{i+1}_{max})^{\frac{1}{2}} \cdot \frac{\partial \{\inp{i+1}\}_{d_{i+1}}}{\partial \{\inp{i}\}_{d_i} } \right] \Biggr\}^2\\
    &= \maxg{i}_{max} \left[\sum_{d_{i+1}}  \frac{\partial \{\inp{i+1}\}_{d_{i+1}}}{\partial \{\inp{i}\}_{d_i} } \right]^2.
\end{align*}
The squared term can be further bounded by using the chain rule, namely, 
\begin{align*}
    \{\maxg{i}\}_{k,d_i} &\leq \maxg{i+1}_{max} \left[\sum_{d_{i+1}} \sign{\{\inp{i+1}\}_{d_{i+1}}}\cdot W^{(i)}_{d_i, d_{i+1}}\right]^2\\
    &\leq \maxg{i+1}_{max} \left[\sum_{d_{i+1}} |W^{(i)}_{d_i, d_{i+1}}|\right]^2.
\end{align*}
Recall that our goal is to bound $\maxg{i}_{max}$. By definition, we take the max on both sides over $d_i$ and $k$ (although the class dimension is already gone). The direction of the inequality still holds because quantities on both sides are all non-negative.
\begin{align*}
     \maxg{i}_{max}  &\leq \maxg{i+1}_{max}  \max_{d_i}\left[\sum_{d_{i+1}} |W^{(i)}_{d_i, d_{i+1}}|\right]^2.
\end{align*}
The order of max and square can be exchanged because $|W^{(i)}_{d_i, d_{i+1}}|$ is non-negative. Thus,
\begin{align*}
     \maxg{i}_{max}  &\leq \maxg{i+1}_{max}  \left[\max_{d_i}\sum_{d_{i+1}} |W^{(i)}_{d_i, d_{i+1}}|\right]^2.
\end{align*}
The last term inside the square is the definition of the $\ell_1$ operator norm so we directly write 
\begin{align*}
    \maxg{i}_{max} \leq \maxg{i+1}_{max} \cdot ||W^{(i)}||^2_1.
\end{align*}

\subsection{Derivations of Propositions}
\paragraph{Proposition~\ref{prop:at_trh}} Given a training dataset $D^m$ and the adversarial input example $x'$ for each example $x$, the top-layer TrH of the AT loss (Definition \ref{def:at}) is equal to
\begin{align}
\Tr(\triangledown^2_{\theta_t} \atloss{\theta}) = \frac{1}{m} \sum_{(x, y) \in D^m} \TrH_{\texttt{AT}}(x';\theta), \nonumber\\
\text{where }\;\; \TrH_{\texttt{AT}}(x';\theta)= ||f_{\theta_b}(x')||^2_2 \cdot \mathbf{1}^\top h(x', \theta) \nonumber.
\end{align} 
\textit{Proof.} We firstly write the expression of $\atloss{\theta}$ based on Definition~\ref{def:at},
\begin{align*}
  \atloss{\theta} = \frac{1}{m} \sum_{(x,y) \in D^m} \max_{||\epsilon||_p \leq \delta}\CE((x + \epsilon, y), F_\theta).
\end{align*}
Let $x'$ be an adversarial example, namely 
\begin{align*}
    x' = \arg\max_{||\epsilon||_p \leq \delta}\left[\CE((x, y), F_\theta)\right] + x.
\end{align*}
AT loss can be simplified as 
\begin{align*}
  \atloss{\theta} = \frac{1}{m} \sum_{(x, y) \in D^{m}} \CE((x', y), F_\theta) =  \frac{1}{m} \sum_{(x, y) \in D^{m}} \sum_{k} - y_{k} \log s(g'_{k}),
\end{align*} 
where $g'_{k} = \{\theta^\top_t f_{\theta_b}(x')\}_k$ is the logit score of class $k$ at the adversarial example $x'$ and $s(\cdot)$ is the softmax function. Let's consider the loss at each adversarial point
\begin{align*}
    R = \sum_{k} - y_{k} \log s(g'_{k}). 
\end{align*}
We want to compute the derivatives of $R$ with respect to the weight at the top layer  $\{\theta_t\}_{jk}$. For the ease of the notation, let's define $w=\theta_t$ so $w_{jk} = \{\theta_t\}_{jk}$ and $\triangledown_{w_{jk}}R$ is equal to
\begin{align}
    \triangledown_{w_{jk}}R  = \sum_{k} - y_{k}  \triangledown_{w_{jk}}[\log s(g'_{k})] =  \{f_{\theta_b}(x')\}_j(s(g'_{k}) - y_{k}). \label{eq:grad_of_ce}
\end{align}
The transition to Eq.~\ref{eq:grad_of_ce} uses the standard form of the gradient of Cross Entropy loss with the softmax activation. To compute the trace of a Hessian, we can skip the cross terms in Hessian (e.g. $\triangledown_{w_{j_1k}, w_{j_2k}}R$) because they are not on the diagonal of the matrix. Thus, we are only interested in $\triangledown^2_{w_{jk}}R$, which are
\begin{align*}
\triangledown^2_{w_{jk}}R&=  \triangledown_{w_{jk}} \left[\triangledown_{w_{jk}}R\right] = \triangledown_{w_{jk}} \left[ x'_{j}(s(g'_{k}) - y_{k})\right] \\
    &= \{f_{\theta_b}(x')\}_j \cdot \{f_{\theta_b}(x')\}_j \cdot s(g'_k)(1 - s(g'_k))\\
    &= \{f_{\theta_b}(x')\}^2_j (s(g'_k) - s^2(g'_k)).
\end{align*}
Eventually, to compute the trace of Hessian matrix we sum all diagonal terms so that 
\begin{align*}
\Tr(\triangledown^2_{w_{jk}}R) = \sum_{j, k} \triangledown^2_{w_{jk}}R = \sum_{j, k} \{f_{\theta_b}(x')\}^2_j (s(g'_k) - s^2(g'_k)) = ||f_{\theta_b}(x')||^2_2\cdot \mathbf{1}^\top h(x', \theta)
\end{align*}
and we denote $\Tr(\triangledown^2_{w_{jk}}R)$ as $\TrH_{\texttt{AT}}(x';\theta)$ to complete the proof.

\paragraph{Proposition~\ref{prop:trades_trh}} Under the same assumption as in Proposition~\ref{prop:at_trh}, the top-layer TrH of the TRADES loss (Def' \ref{def:trades}) is equal to
\begin{align*}
    &\Tr(\triangledown^2_{\theta_t} \tradesloss{\theta}) = \frac{1}{m} \sum_{(x, y) \in D^m} \TrH_{\texttt{T}}(x, x'; \lambda_t, \theta), \nonumber\\
&\text{where,}\;\; \TrH_{\texttt{T}}(x, x'; \lambda_t, \theta) = ||f_{\theta_b}(x)||^2_2 \cdot \mathbf{1}^\top h(x, \theta)  +\lambda_t ||f_{\theta_b}(x')||^2_2 \cdot \mathbf{1}^\top h(x', \theta).
\end{align*}

Before we start our proof of this proposition, we introduce the following useful lemmas.

\begin{lemma}\label{lemma:softmax_grad}
    Let $s(g)$ be the softmax output of the logit output $g$ computed at input x, then 
    \begin{align}
        \frac{\partial s(g)}{\partial g} = \Jprob =  diag[s(g)] - s(g)\cdot\{s(g)\}^\top
    \end{align}
    where $diag(v)$ returns an identity matrix with its diagonal replaced by a vector $v$. 
\end{lemma}
\textit{Proof.} 
\begin{align*}
     \frac{\partial s(g_i)}{\partial g_k} = s(g_i)(\mathbb{I}\left[i = k\right] - s(g_k)), \Jprob_{ik} = s(g_i)\cdot\mathbb{I}\left[i = k\right] - s(g_i)s(g_k) \implies \frac{\partial s(g)}{\partial g} = \Jprob.
\end{align*}

\begin{lemma}\label{lemma:log_softmax_grad}
    Let $\log s(g(x))$ be the log softmax output of the logit output g computed at input x, then 
    \begin{align*}
        \frac{\partial 
        \log s(g(x))}{\partial g(x)} = \Jlprob =  I - \mathbf{1}\cdot\{s(g(x))\}^\top
    \end{align*}
    where $I$ is the identity matrix. 
\end{lemma}

\textit{Proof.} 
\begin{align*}
     \frac{\partial \log s(g_i)}{\partial g_k} = \mathbb{I}\left[i = k\right] - s(g_k), \Jlprob_{ik} = \mathbb{I}\left[i = k\right] - s(g_k) \implies \frac{\partial s(g)}{\partial g} = \Jlprob.
\end{align*}

Now we are ready to present our proof of the Proposition~\ref{prop:trades_trh}.

\textit{Proof.} We write the expression of $\tradesloss{\theta}$ based on as Definition~\ref{def:trades}.
\begin{align*}
    \tradesloss{\theta} &\defeq \frac{1}{m} \sum_{(x,y)\in D^m} \Big[ \CE((x, y), g_\theta) + \lambda_t \cdot \max_{||\epsilon||_p \leq \delta}\KLL((x,x+\epsilon), g_\theta) \Big],
\end{align*}
where $\KLL((x, x+\epsilon), g_\theta) = \KL(s(g_\theta(x)) || s(g_\theta(x+\epsilon))$. Thus, we can compute the trace of Hessian on the top layer for the CE loss and the KL loss, respectively. Namely, we derive $\triangledown^2_{\theta_t} [\CE((x, y), F_\theta)]$ and $\triangledown^2 [\max_{||\epsilon||_p \leq \delta}\KLL((x, x+\epsilon), F_\theta)]$.

First, we see that the expression of $\triangledown^2_{\theta_t}[\CE((x, y), F_\theta)]$ is similar to the result in Proposition~\ref{prop:at_trh} by replacing the adversarial input with the clean input. With this similarity, we directly write 
\begin{align}
    \Tr\{\triangledown^2_{\theta_t}[\CE((x, y), F_\theta)]\} = ||f_{\theta_b}(x)||^2_2\cdot \mathbf{1}^\top h(x, \theta). \label{eq:Tr-of-CE-in-TRADES}
\end{align}
Second, by denoting
\begin{align*}
    x' = \arg\max_{||\epsilon||_p \leq \delta}\left[\KLL((x, x+\epsilon), F_\theta)\right] + x,
\end{align*}
the rest of the proof now focuses on deriving $\triangledown^2_{\theta_t}[\KLL((x, x'), F_\theta)]$ where 
\begin{align*}
    \KLL((x, x'), F_\theta) = -\sum_i s(g_{i}) \log(s(g'_{i})) + \left[ \sum_i s(g_{i}) \log(s(g_{i}))\right], \quad g_i = \{\theta^\top_t f_{\theta_b}(x)\}_i, g'_i =\{ \theta^\top_t f_{\theta_b}(x')\}_i.
\end{align*}
For the ease of the notation, let $w \defeq \theta_t$ so $w_{jk} \defeq \{\theta_t\}_{jk}$ and let $K \defeq  \KLL((x, x'), F_\theta)$. 
To find $\triangledown^2_{\theta_t}K$, we first write the first-order derivative of $K$ with respect to $w$,
\begin{align}
    \frac{\partial K}{\partial w_{jk}} = -\sum_i s(g_i) \frac{\partial }{\partial g'_k}\left[\log(s(g'_{i}))\right] \frac{\partial g'_k}{\partial \partial w_{jk}} + \underbrace{\sum_i \frac{\partial K}{\partial s(g_i)}\frac{\partial s(g_i)}{\partial g_k} \frac{\partial g_k}{\partial \partial w_{jk}}}_{\text{gradient through } g_k}. \label{eq:KLL_1st_grad}
\end{align}
Next, we discuss two cases depending on whether or not we stop gradient on $g$ (the logit output of the clean input) in Eq.~\ref{eq:KLL_1st_grad}. 
In Case I where the gradient on $g$ is stopped, the second term of Eq.~\ref{eq:KLL_1st_grad} will vanish. This leads to a simpler but practically more stable objective function. 

\paragraph{Case I: Stop Gradient on $g$.} In this case, 
\begin{align}
\frac{\partial K}{\partial w_{jk}} &= -\sum_is(g_i) \frac{\partial }{\partial g'_k}\left[\log(s(g'_{i}))\right] \frac{\partial g'_k}{\partial \partial w_{jk}} \nonumber\\
&=\{f_{\theta_b}(x')\}_j (s(g'_k) - s(g_k)).\label{eq:KLL_1st_grad_on_adv_logit}
\end{align} 
By comparing Eq.~\ref{eq:KLL_1st_grad_on_adv_logit} with Eq.~\ref{eq:grad_of_ce} and treating $s(g_k)$ as constants, we can quickly write out the second-order derivative as 
\begin{align*}
    \frac{\partial^2 K}{\partial w^2_{jk}} = \{f_{\theta_b}(x')\}^2_j (s(g'_k) - s^2(g'_k)).
\end{align*}
Recalling $h(x',\theta) = s(g') - s^2(g')$, therefore,
\begin{align}
    \Tr\{\triangledown^2_{\theta_t}[\KLL((x, x'), F_\theta)]\} = \Tr\{\triangledown^2_{\theta_t}K\} =\sum_{jk} \frac{\partial^2 K}{\partial w^2_{jk}} = ||f_{\theta_b}(x')||^2_2 \cdot \mathbf{1}^\top h(x', 
    \theta).\label{eq:Case-I-Tr-of-KLL-in-TRADES}
\end{align}
Finally, we combine Eq.~\ref{eq:Tr-of-CE-in-TRADES} and~\ref{eq:Case-I-Tr-of-KLL-in-TRADES} to arrive at
\begin{align}
    \Tr\Big\{\triangledown^2_{\theta_t}\Big[ \CE((x, y), g_\theta) + \lambda_t \cdot \max_{||\epsilon||_p \leq \delta}&\KLL((x,x+\epsilon), g_\theta) \Big]\Big\} = ||f_{\theta_b}(x)||^2_2 \cdot \mathbf{1}^\top h(x, \theta)  +\lambda_t ||f_{\theta_b}(x')||^2_2 \cdot \mathbf{1}^\top h(x', \theta),\nonumber\\
    &\text{where }x' = \arg\max_{||\epsilon||_p \leq \delta}\left[\KLL((x, x+\epsilon), F_\theta)\right] + x\nonumber
\end{align}
By denoting $\TrH_{\texttt{t}}(x, x'; \lambda_t, \theta) \defeq \Tr\Big\{\triangledown^2_{\theta_t}\Big[ \CE((x, y), g_\theta) + \lambda_t \cdot \max_{||\epsilon||_p \leq \delta}\KLL((x,x+\epsilon), g_\theta) \Big]\Big\}$, we complete the proof for this case and this is the statement shown in Proposition~\ref{prop:trades_trh}.

\paragraph{Case II: With Gradient on $g$.} We restart our derivation from Eq.~\ref{eq:KLL_1st_grad} and expand the first term as follows
\begin{align*}
    \frac{\partial K}{\partial w_{jk}} = \{f_{\theta_b}(x')\}_j (s(g'_k) - s(g_k)) + \sum_i \frac{\partial K}{\partial s(g_i)}\frac{\partial s(g_i)}{\partial g_k} \frac{\partial g_k}{\partial \partial w_{jk}}. 
\end{align*}
Here
\begin{align*}
    \sum_i \frac{\partial K}{\partial s(g_i)}\frac{\partial s(g_i)}{\partial g_k} \frac{\partial g_k}{\partial \partial w_{jk}} = -\{f_{\theta_b}(x)\}_j\sum_i \frac{\partial s(g_i)}{\partial g_k}\log s(g'_i) + \{f_{\theta_b}(x)\}_j \sum_i \left[\frac{\partial s(g_i)}{\partial g_k}\log s(g_i) +s(g_i) \frac{\partial \log s(g_i)}{\partial g_k} \right]
\end{align*}
Using Lemma~\ref{lemma:softmax_grad} and~\ref{lemma:log_softmax_grad}, we write
\begin{align*}
    \frac{\partial s(g_i)}{\partial g_k} = \Jprob_{ik}, \quad \quad\frac{\partial \log s(g_i)}{\partial g_k} = \Jlprob_{ik}, \quad  \Jprob_{ik} = s(g_i)\Jlprob_{ik}
\end{align*}
and 
\begin{align*}
    \sum_i \frac{\partial K}{\partial s(g_i)}\frac{\partial s(g_i)}{\partial g_k} \frac{\partial g_k}{\partial \partial w_{jk}} = -\{f_{\theta_b}(x)\}_j\sum_i \Jprob_{ik}\log s(g'_i) + \{f_{\theta_b}(x)\}_j \sum_i \left[\Jprob_{ik}\log s(g_i) +\Jprob_{ik} \right].
\end{align*}
Therefore, the first-order derivative of $K$ with respect to $w_{jk}$ is 
\begin{align*}
    \frac{\partial K}{\partial w_{jk}} &= \{f_{\theta_b}(x')\}_j (s(g'_k) - s(g_k)) -\{f_{\theta_b}(x)\}_j\sum_i \Jprob_{ik}\log s(g'_i) + \{f_{\theta_b}(x)\}_j \sum_i \left[\Jprob_{ik}\log s(g_i) +\Jprob_{ik} \right]\\
    &=\underbrace{\{f_{\theta_b}(x')\}_js(g'_k)}_{K'} \underbrace{- \{f_{\theta_b}(x')\}_js(g_k)}_{K_1}\underbrace{-\{f_{\theta_b}(x)\}_j\sum_i \Jprob_{ik}\log s(g'_i)}_{K_2} + \underbrace{\{f_{\theta_b}(x)\}_j \sum_i \left[\Jprob_{ik}\log s(g_i) +\Jprob_{ik} \right]}_{K_3}.
\end{align*}
To calculate the seconder-order derivative of $K$ w.r.t $w_{jk}$, we find the derivative of $K', K_1, K_2, K_3$ w.r.t $w_{jk}$, respectively.

\paragraph{(1) Derivative of $K'$.} $K'$ is simply the result we have already obtained in Case I (see Eq.~\ref{eq:KLL_1st_grad_on_adv_logit}); therefore, 
\begin{align*}
    \frac{\partial K'}{\partial w_{jk}} = \{f_{\theta_b}(x')\}^2_j (s(g'_k) - s^2(g'_k)).
\end{align*}
Thus,
\begin{align*}
    \Tr(\frac{\partial K'}{\partial w_{jk}}) = ||f_{\theta_b}(x')||^2_2 \cdot \mathbf{1}^\top h(x', 
    \theta).
\end{align*}
\paragraph{(2) Derivative of $K_1$.}
\begin{align*}
    \frac{\partial K_1}{\partial w_{jk}} &= -\{f_{\theta_b}(x')\}_j \{f_{\theta_b}(x)\}_j \frac{\partial s(g_k)}{\partial g_{k}} \\
    &= -\{f_{\theta_b}(x')\}_j \{f_{\theta_b}(x)\}_j \Jprob_{kk} \\
    &= -\{f_{\theta_b}(x')\}_j \{f_{\theta_b}(x)\}_j (s(g_k) - s^2(g_k)).
\end{align*}
Thus,
\begin{align*}
    \Tr(\frac{\partial K_1}{\partial w_{jk}}) &= -\sum_{jk}\{f_{\theta_b}(x')\}_j \{f_{\theta_b}(x)\}_j (s(g_k) - s^2(g_k))\\
    &=-f_{\theta_b}(x')^\top f_{\theta_b}(x)\cdot \mathbf{1}^\top h(x, 
    \theta)
\end{align*}

\paragraph{(3) Derivative of $K_2$.}
\begin{align*}
    \frac{\partial K_2}{\partial w_{jk}} &= -\{f_{\theta_b}(x)\}_j\sum_i \left[\frac{\partial \Jprob_{ik}}{\partial g_{k}}\frac{\partial g_{k}}{\partial w_{jk}} \log s(g'_i)  + \Jprob_{ik} \frac{\partial \log s(g'_i)}{\partial g'_k} \frac{\partial g'_k}{\partial w_{jk}} \right]\\
    &=-\{f_{\theta_b}(x)\}_j\sum_i \left[\frac{\partial \Jprob_{ik}}{\partial g_{k}} \{f_{\theta_b}(x)\}_j \log s(g'_i)  + \Jprob_{ik} \Jlprob'_{ik} \{f_{\theta_b}(x')\}_j \right]
\end{align*}
Notice that
\begin{align*}
    &i = k \implies \frac{\partial \Jprob_{ik}}{\partial g_{k}} =\frac{\partial \Jprob_{kk}}{\partial g_{k}} = \frac{\partial}{\partial g_{k}}(s(g_k) - s^2(g_k)) = \Jprob_{kk} - 2s(g_k)\Jprob_{kk};\\
    \text{and }&i \neq k \implies \frac{\partial \Jprob_{ik}}{\partial g_{k}} = \frac{\partial}{\partial g_{k}}(- s(g_i)s(g_k)) = -2s(g_i)\Jprob_{kk}.
\end{align*}
Thus, 
\begin{align*}
    \frac{\partial \Jprob_{ik}}{\partial g_{k}} = \Jprob_{kk}( \mathbb{I}[k=i] - s(g_i)) = \Jprob_{kk} \{\Jlprob^\top\}_{ik} =\Jprob_{kk} \Jlprob_{ki}.
\end{align*}
As a result,
\begin{align*}
    \frac{\partial K_2}{\partial w_{jk}}&=-\{f_{\theta_b}(x)\}_j\sum_i \left[ \Jprob_{kk}  \Jlprob_{ki}\{f_{\theta_b}(x)\}_j \log s(g'_i)  + \Jprob_{ik} \Jlprob'_{ik} \{f_{\theta_b}(x')\}_j \right]\\
    &=-\{f_{\theta_b}(x)\}^2_j \Jprob_{kk} \sum_{i}\Jlprob_{ki} \log s(g'_i) - \{f_{\theta_b}(x)\}_j \{f_{\theta_b}(x')\}_j\sum_i\Jprob_{ik} \Jlprob'_{ik}\\
    &=(-\{f_{\theta_b}(x)\}^2_j \Jprob_{kk})(\Jlprob_{k}\cdot\log s(g')) - \{f_{\theta_b}(x)\}_j \{f_{\theta_b}(x')\}_j(\{\Jprob^\top\}_{k} \cdot \{\Jlprob'^{\top}\}_{k}).
\end{align*}
Thus,
\begin{align*}
    \Tr(\frac{\partial K_2}{\partial w_{jk}}) &= \sum_{jk}\left[(-\{f_{\theta_b}(x)\}^2_j \Jprob_{kk})(\Jlprob_{k}\cdot\log s(g')) - \{f_{\theta_b}(x)\}_j \{f_{\theta_b}(x')\}_j(\{\Jprob^\top\}_{k} \cdot \{\Jlprob'^{\top}\}_{k})\right]\\
    &=-||f_{\theta_b}(x)||^2_2\cdot \psi^{'\top} h(x, \theta) - f^\top_{\theta_b}(x')f_{\theta_b}(x)\cdot \mathbf{1}^\top \omega',
\end{align*} where $\psi', \omega'$ are vectors such that $\psi'_k = \Jlprob_{k}\cdot\log s(g'), \omega'_k = \{\Jprob^\top\}_{k} \cdot \{\Jlprob'^{\top}\}_{k}$.

\paragraph{(4) Derivative of $K_3$.}
\begin{align*}
    \frac{\partial K_3}{\partial w_{jk}} &= \{f_{\theta_b}(x)\}_j \sum_i\left[ \frac{\partial \Jprob_{ik}}{\partial g_k} \frac{\partial g_k}{\partial w_{jk}}\log s(g_i) + \Jprob_{ik} \frac{\partial \log s(g_i)}{\partial g_k}\frac{\partial g_k}{\partial w_{jk}} + \frac{\partial \Jprob_{ik}}{\partial g_k} \frac{\partial g_k}{\partial w_{jk}} \right]\\
    &= \{f_{\theta_b}(x)\}^2_j \sum_i\left[\Jprob_{kk} \Jlprob_{ki}\log s(g_i) + \Jprob_{ik} \Jlprob_{ik}  +\Jprob_{kk} \Jlprob_{ki}\right]\\
    &=\{f_{\theta_b}(x)\}^2_j\left[ \Jprob_{kk} (\Jlprob_{k} \cdot \log s(g)) + (\{\Jprob^\top\}_{k} \cdot \{\Jlprob^{\top}\}_{k}) +\Jprob_{kk}(\underbrace{\Jlprob_k\cdot\mathbf{1})}_{\text{row sum is 0}}\right]\\
    &=\{f_{\theta_b}(x)\}^2_j\left[ \Jprob_{kk} (\Jlprob_{k} \cdot \log s(g)) + (\{\Jprob^\top\}_{k} \cdot \{\Jlprob^{\top}\}_{k}) \right].
\end{align*}
Thus,
\begin{align*}
    \Tr(\frac{\partial K_3}{\partial w_{jk}}) &= \sum_{jk}\left[\{f_{\theta_b}(x)\}^2_j\left[ \Jprob_{kk} (\Jlprob_{k} \cdot \log s(g)) + (\{\Jprob^\top\}_{k} \cdot \{\Jlprob^{\top}\}_{k}) \right]\right]\\
    &=||f_{\theta_b}(x)||^2_2\cdot (\psi^\top h(x, \theta) + \mathbf{1}^\top \omega).
\end{align*}
Finally, we have
\begin{align*}
    \Tr(\frac{\partial^2 K}{\partial w^2_{jk}}) &= \Tr\left[\frac{\partial K'}{\partial w_{jk}} + \frac{\partial K_1}{\partial w_{jk}} + \frac{\partial K_2}{\partial w_{jk}} + \frac{\partial K_3}{\partial w_{jk}}\right]\\
    &= ||f_{\theta_b}(x')||^2_2 \cdot \mathbf{1}^\top h(x', 
    \theta) + G(x, x';\theta)
\end{align*}where
\begin{align*}
G(x, x';\theta) &= -f_{\theta_b}(x')^\top f_{\theta_b}(x)\cdot \mathbf{1}^\top h(x, 
    \theta) -f_{\theta_b}(x')^\top f_{\theta_b}(x)\cdot \mathbf{1}^\top h(x, 
    \theta) -||f_{\theta_b}(x)||^2_2\cdot \psi^{'\top} h(x, \theta) \\
    &- f^\top_{\theta_b}(x')f_{\theta_b}(x)\cdot \mathbf{1}^\top \omega' + ||f_{\theta_b}(x)||^2_2\cdot (\psi^\top h(x, \theta) + \mathbf{1}^\top \omega).
\end{align*}
We arrive at
\begin{align}
    \Tr\Big\{\triangledown^2_{\theta_t}\Big[ \CE((x, y), g_\theta) &+ \lambda_t \cdot \max_{||\epsilon||_p \leq \delta}\KLL((x,x+\epsilon), g_\theta) \Big]\Big\} \\
    &= ||f_{\theta_b}(x)||^2_2 \cdot \mathbf{1}^\top h(x, \theta)+\lambda_t( ||f_{\theta_b}(x')||^2_2 \cdot \mathbf{1}^\top h(x', \theta) + G(x, x';\theta)),\nonumber
\end{align}where
\begin{align*}
    x' = \arg\max_{||\epsilon||_p \leq \delta}\left[\KLL((x, x+\epsilon), F_\theta)\right] + x\nonumber
\end{align*}
Finally, we complete the derivation by denoting $\TrH_{\texttt{t}}(x, x'; \lambda_t, \theta) \defeq \Tr\Big\{\triangledown^2_{\theta_t}\Big[ \CE((x, y), g_\theta) + \lambda_t \cdot \max_{||\epsilon||_p \leq \delta}\KLL((x,x+\epsilon), g_\theta) \Big]\Big\}$.

\section{Impact of Top-Layer Regularization}\label{appendix:efficiency-example}
\begin{figure}[t]
    \centering
\includegraphics[width=0.4\textwidth]{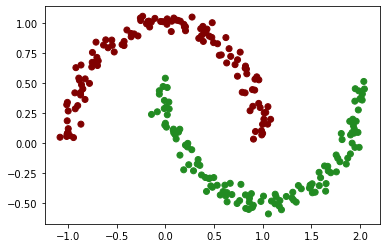}
    \caption{Visualization of Two Moons dataset. The task is to classify the points into two classes (red and green).}
    \label{fig:two-moons}
\end{figure}

In this section, we provide a full description of Example~\ref{eg:toy_network}, as well as some additional results and analysis.
\begin{itemize}
    \item \textbf{Dataset.} We use the Two Moons Dataset where the input features $x_i \in \mathbb{R}^2$ and the label $y_i \in \{0, 1\}$ (Figure~\ref{fig:two-moons}).
    \item \textbf{Network.} We use a dense network Dense(100)-ReLU-Dense(100)-ReLU-Dense(2). 
    \item \textbf{Training without Regularization (Standard).} We train the network with AT(\base), such that it is robust in an $\ell_\infty$ ball of size 0.02 with 1 PGD step. We use a momentum-SGD with learning rate 0.1 for 100 epochs. 
    \item \textbf{Training with Top TrH Regularization (Top).} We further add the TrH of the top layer as a regularization term in the loss. We compute the TrH at the top layer using Proposition~\ref{prop:at_trh}. The regularizing coefficient for the top-layer TrH is 0.5 (i.e. $loss = AT + 0.5 * \TrH_{top}$) as we find it needs a stronger penalty due to the lack of TrH from other layers. 
    \item \textbf{Training with Full TrH Regularization (Full).} In additional to the Standard setup, we add the TrH of the full network as a regularization term in the loss. We numerically compute the Hessian and its trace. The regularizing coefficient for the full-network TrH is 0.05 (i.e. $loss = AT + 0.05 * \TrH_{full}$). 
\end{itemize}

\begin{figure}[t]
    \centering
\includegraphics[width=\textwidth]{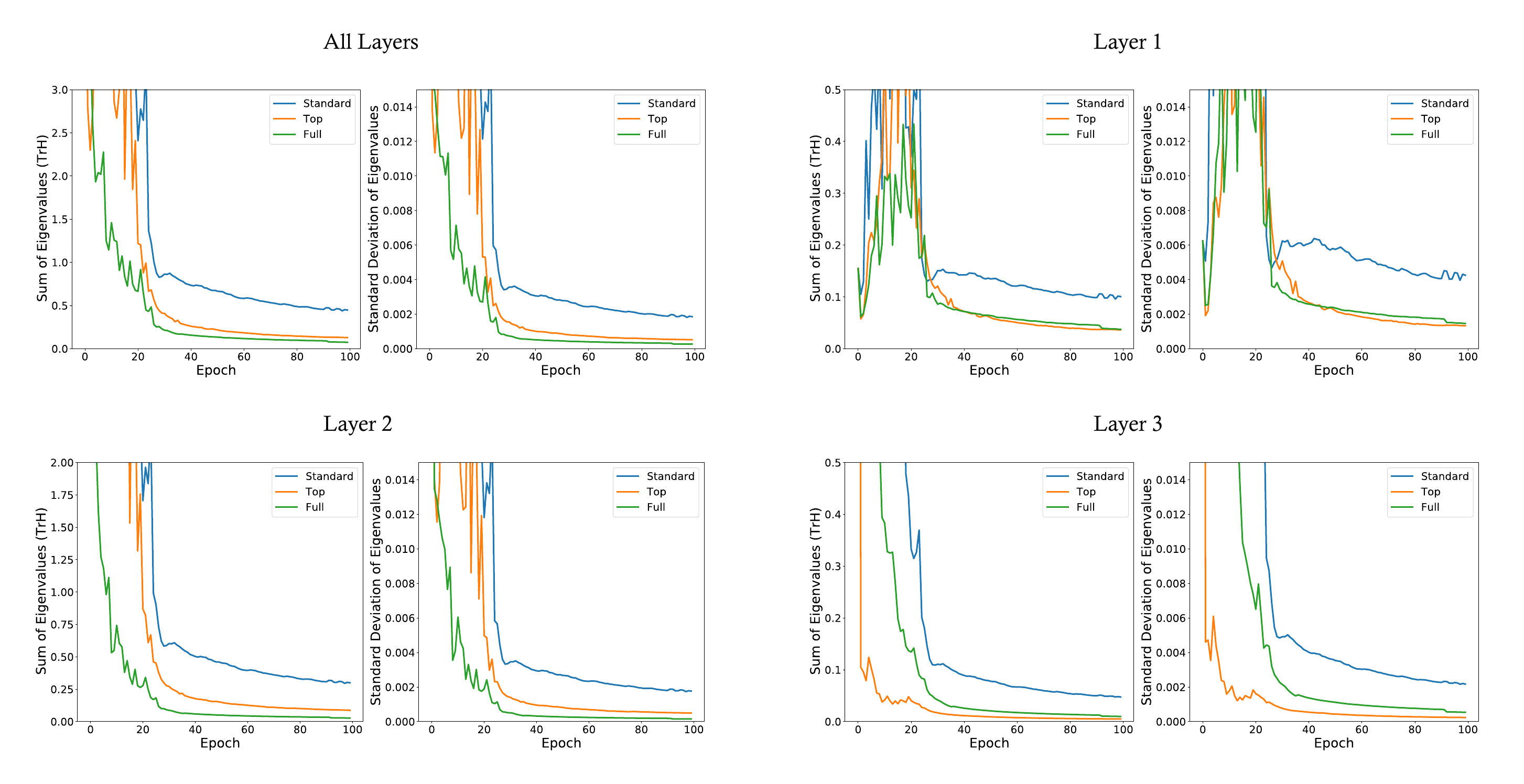}
    \caption{The figure shows pairs of plots for the sum and the standard deviation of the Hessian matrix eigenvalues. The top-left pair corresponds to the Hessian of all layers, while the rest to each of the three layers separately, with Layer 3 being the top layer (note that the sum of eigenvalues is exactly the trace of Hessian). 
    As can be seen on the top-left , the sum and standard deviation decrease in standard training (blue) and moreover, this effect is amplified by direct TrH regularization with similar results for full network regularization (green) and top-layer regularization (orange). This similarity can be explained by our Thm.\ref{thm:grad-prop}}
    \label{fig:eigen-spectrum}
\end{figure}

Figure~\ref{fig:eigen-spectrum}, shows pairs of plots of  \textbf{the sum and standard deviation of the eigenvalues of the Hessian matrix} across  all layers, as well as for each of the three layers separately. 
We compare the three setting (1) no regularization (Standard, blue); (2) the top-layer TrH regularization (Top, orange); and (3) the full-network TrH regularization (Full, green). 

Focusing on the top-left plot in Figure~\ref{fig:eigen-spectrum}, we see that the TrH decreases across time with standard training (blue), but at the same time, that  full-network and top-layer TrH regularization decrease it even further. 
In particular, top-layer TrH regularization is nearly as effective as directly regularizing the TrH for all layers. 
The standard deviation plots (on the right side of each TrH plot) show a a decrease in the standard deviation, which implies there is a contraction effect on the eigenvalues of the Hessian, with both positive and negative values approaching towards 0. This rules out the possibility that the TrH reduction comes from an increase in the magnitude of the negative eigenvalues. In fact, TrH regularization effectively contracts the magnitude of all eigenvalues and leads to a smoother region in the loss surface. 

We further plot the TrH and standard deviation of Hessian eigenvalues for each individual layer respectively (where Layer 3 is the top layer). 
Except for the first layer, whose eigenvalues seem to be small from the onset, the eigenvalue contraction effect is evident as training progresses, with a stronger and similar effect for both TrH regularization settings.

\begin{table*}[b]
    \centering
    \small
    \scalebox{0.95}{\begin{tabular}{llll}
     \multicolumn{4}{c}{\textbf{Pre-Selected Hyper-parameters (CIFAR-10/100)}}\\
    \toprule
    \toprule
    Parameter& \multicolumn{2}{l}{Reason To Select} & Final Frozen Value\\
    \midrule
     \texttt{img\_size}& \multicolumn{2}{l}{to be compatible with $\epsilon$} & $32\times 32$\\
     \texttt{patch\_size}& \multicolumn{2}{l}{same as \cite{Mahmood_2021_ICCV}} & $4\times 4$\\
     \texttt{parameter\_init}& \multicolumn{2}{l}{publicly available} & ImageNet21K checkpoint\\
     \texttt{batch\_size}& \multicolumn{2}{l}{memory} & 768\\
     \texttt{warm\_up iterations}& \multicolumn{2}{l}{standard} & 500\\
    \midrule
    \multicolumn{4}{c}{\textbf{Pre-tuning Stage}}\\
    \toprule
    \toprule
     Parameter& Range & Explanation & Final Frozen Value\\
     \midrule
     \texttt{optimizer} & \{`sgd+momentum', `adam'\}& optimizer & sgd+momentum  \\
     \texttt{base\_lr} & $\{0.001,0.01, 0.1\}$& initial learning rate & 0.1\\
     \texttt{l2\_reg} & $\{0, 0.0001, 0.001\}$ & coefficient for L2 regularization & 0 \\
     \texttt{data\_aug} & \{`fb', `crop'\} & data augmentation method & `fb'\\
     & &\quad `fb': flip and random brightness\\
     & &\quad `crop': randomly crop and upsample\\
     \texttt{downsample} & \{`cubic', `nearest', `bilinear' \} & downsample method for the first kernel to & cubic \\
     & &\quad fit the patch size from $16\times 16$ to $4\times 4$~\cite{Mahmood_2021_ICCV}&\\
     \texttt{patch\_stride}&$\{2, 4\}$ & the stride to create image patches  & 2\\
     \texttt{data\_range} & \{$[-1,1], [0, 1]$, `centered'\} &data range & centered \\
    & &`centered': 0-mean and 1-std & \\
\texttt{use\_cutmix}& \{True, False\} & whether to use cutmix augmentation &False\\
    \bottomrule
    \end{tabular}}
    \caption{Frozen Hyper-parameters for CIFAR-10 and CIFAR-100 (including DDPM images) in All Experiments.}
    \label{tab:pre-tuning for CIFAR-10 and CIFAR-100}
\end{table*}

\begin{table*}[t]
    \centering
    \small
    \scalebox{0.95}{\begin{tabular}{llll}
     \multicolumn{4}{c}{\textbf{Pre-Selected Hyper-parameters (ImageNet)}}\\
    \toprule
    \toprule
    Parameter& \multicolumn{2}{l}{Reason To Select} & Final Frozen Value\\
    \midrule
     \texttt{data\_range}& \multicolumn{2}{l}{to align with ImageNet21K checkpoint} & $[-1, 1]$\\
     \texttt{img\_size}& \multicolumn{2}{l}{to be compatible with $\epsilon$} & $224\times 224$\\
     \texttt{patch\_size}& \multicolumn{2}{l}{standard} & $16\times 16$\\
     \texttt{parameter\_init}& \multicolumn{2}{l}{publicly available} & ImageNet21K checkpoint\\
     \texttt{batch\_size}& \multicolumn{2}{l}{memory} & 64\\
     \texttt{warm\_up iterations}& \multicolumn{2}{l}{standard} & 500\\
    \midrule
    \multicolumn{4}{c}{\textbf{Pre-tuning Stage}}\\
    \toprule
    \toprule
     Parameter& Range & Explanation & Final Frozen Value\\
     \midrule
     \texttt{optimizer} & \{`sgd+momentum', `adam'\}& optimizer & `sgd+momentum'  \\
     \texttt{base\_lr} & $\{0.001, 0.01, 0.1\}$& initial learning rate & $0.01$  \\
     \texttt{decay\_type} & \{`multistep', `cosine'\}& the function used to schedule lr decay & cosine  \\
     \texttt{l2\_reg} & $[0.0001, 0.001]$ & coefficient for $\ell_2$ regularization & $0.0001$\\
     \texttt{data\_aug} & \{`fb', `crop'\} & data augmentation method & `fb'\\
     & &\quad `fb': flip and random brightness\\
     & &\quad `crop': randomly crop and upsample\\
\texttt{use\_cutmix}& \{True, False\} & whether to use cutmix augmentation &False\\
         \bottomrule
    \end{tabular}}
    \caption{Frozen Hyper-parameters for ViT-B16 and ViT-L16 to reproduce results in Table.~\ref{tab:main_results}.}
    \label{tab:pre-tuning for imagenet}
\end{table*}

\section{Hyper-parameters Shared by All Methods}\label{appendix:pre-tuning}
\paragraph{Pre-Tuning.} Training ViTs can sometimes be challenging due to a large amount of hyper-parameters. For the choice of the parameters that are shared across different defense methods, e.g. batch size, patch size, training iterations, and etc., we do a large grid search and choose the parameter setting that produce the best results on TRADES(\base) and use it for all the methods. This step is referred to as pre-tuning and is done per-dataset. 

\paragraph{Pre-selected Hyper-parameters.} There is a set of hyper-parameters requiring no tuning because they are commonly selected in the literature. In the top of Table~\ref{tab:pre-tuning for imagenet} and~\ref{tab:pre-tuning for CIFAR-10 and CIFAR-100}, we write down the these parameters and explain the reason for choosing a particular value. 

\paragraph{Tunable Hyper-parameters.} In the bottom of Table~\ref{tab:pre-tuning for imagenet} and~\ref{tab:pre-tuning for CIFAR-10 and CIFAR-100}, we show our choice of hyper-parameters for ImageNet and CIFAR-10/100, respectively. This includes
\begin{itemize}
    \item \texttt{optimizer.} We tested a momentum-SGD (\texttt{sgd+momentum}) and an Adam optimizer (\texttt{adam}). We found that momentum-SGD is more stable in fine-tuning the ViT from a pre-trained checkpoint.
    \item \texttt{base\_lr}, \texttt{warm\_up\_iterations} and \texttt{decay\_type}. We linearly increase the learning rate from 0 to the \texttt{base\_lr} during \texttt{warm\_up\_iterations}. After warm-up, we gradually schedule the learning rate based on the \texttt{decay\_type}. We experiment with a multi-step and a cosine decay and find no apparent difference between these two schedulers. In the end, we choose cosine because it has less hyper-parameters to choose compared to the multi-step one.
    \item \texttt{\texttt{l2\_reg}.} On ImageNet, we find $\ell_2$ regularization with a penalty of 0.0001 helps the baseline TRADES(\base). On CIFAR-10/100, we find that $\ell_2$ regularization may not be necessary when using DDPM data. 
    \item \texttt{cutmix.} In both cases we do not find the cut mix augmentation~\cite{yun2019cutmix} help to improve the results. 
    \item \texttt{data\_range.} On CIFAR10/100, we find that centered data, i.e. normalizing the data to have 0 mean and (close to) 1 standard deviation, provides better results than scaling the images to [-1, 1] (the range of data used by the pre-trained checkpoint). We simply subtract the CIFAR images from the average of per-channel mean (0.47) and divide it with the average of per-channel standard deviation (0.25). For ImageNet, we still use [-1, 1] as the data range. Notice that $\eps$ ball needs to be re-scaled for both data ranges describe above. For example, when reporting results on $\epsilon=0.031$ and using centered data, we need to use $\epsilon / 0.25$ as the actual noise bound passed to the attacker. When normalizing the data to $[-1, 1]$, we need to pass $\epsilon / 0.5$ to the attacker.
     \item \texttt{patch\_size}. The size of the image patch of the input sequence to ViT. $16 \times 16$ is the standard size of pre-trained ViT models on ImageNet21K. Thus, when fine-tuning on ImageNet, we use $16 \times 16$. CIFAR images are a lot smaller compared to ImageNet images. As a result, we use a smaller patch size of $4\times 4$ to produce more patches. Using a smaller patch size requires some modifications to ViT architecture and we discuss this in detail in Appendix~\ref{appendix:architecture}.
    \item \texttt{downsample} and \texttt{patch\_stride.} These are particular to CIFAR images. Please refer to Appendix~\ref{appendix:architecture} for detail. 
\end{itemize}

\section{Hyper-parameters for Specific Methods}\label{appendix:fine-tuning}
We fine-tune ViTs after common hyper-parameters are locked after pre-tuning. For all methods, we take 10 PGD steps on CIFAR-10/100 and 7 steps for ImageNet during the training. We report the best results for each method after trying different sets of hyper-parameters. This usually involves method-specific parameters. We elaborate what hyper-paramter is tuned as follows and report the final values used in the experiments in Table~\ref{tab:fine-tuning for cifar10/100} (CIFAR-10/100) and~\ref{tab:fine-tuning for imagenet} (ImageNet).
\begin{itemize}
    \item \textbf{base}: we use $\lambda_t = 6$ for TRADES training and $\lambda_t$ is consistent across all experiments.  
    \item \textbf{SWA}: we use $\alpha=0.995$ so that $\theta_{\texttt{avg}} \leftarrow 0.995 *  \theta_{\texttt{avg}} + 0.005*\theta^{(t+1)} $ as this value is used in the literature~\cite{Gowal2021ImprovingRU}. Therefore, there is no tuning in SWA.
    \item \textbf{S2O}: The only parameter that requires tuning is the penalty $\alpha$ of the second-order statistic in Eq~\ref{eq:s2o}. In the authors' implementation, we find $1-\alpha$ is used to balance the regularization with the robust loss (AT or TRADES). These hyper-parameters are hard-coded in the latest commit f2d037b\footnote{\url{https://github.com/Alexkael/S2O/tree/f2d037b9611f7322783411825099251f7978f54e}} so we directly use their choice of hyper-parameters. Namely, for AT loss, the finally loss in S2O training is set to $$0.9 * AT\_loss + 0.1 * S2O\_loss$$ and for TRADES training the final loss is $$  0.7 * \frac{1}{m} \sum^m_i \CE((x_i, y_i), F_\theta) + 0.3 * S2O\_loss +\frac{\lambda_t}{m}\sum^m_i \max_{||\epsilon_i||_p \leq \delta}\KLL((x_i,\epsilon_i), F_\theta).  \nonumber $$
    \item \textbf{AWP}: The two hyper-parameters in AWP that requires tuning are $\psi$ and $\delta_{awp}$, where $\psi$ is the function to measure the noise added to the weights and $\delta_{awp}$ is the noise budget. We follow the choice made by Wu et al.~\cite{wu2020adversarial} to choose $\psi$ as the layer-wise $\ell_2$ norm function so that the noise added to the weights in each layer should no greater than the $\ell_2$ norm of the weights multiplied by $\delta_{awp}$. Namely, suppose that $\xi^{(i)}$ is the noise added to a weight matrix $W^{(i)}$ at layer $i$, then we project $\xi^{(i)}$ such that $$\frac{||\xi^{(i)} + W^{(i)}||_2}{  ||W^{(i)}||_2} \leq \delta_{awp}. $$
    For the choice of $\delta_{awp}$, we sweep $\delta_{awp}$ over the interval \{0.0001, 0.0005, 0.001, 0.005, 0.01\}.
    \item \textbf{TrH}: The two hyper-parameters in TrH that requires tuning are the $\ell_2$ weight penalty $\gamma$ and the TrH penalty $\lambda$. For $\gamma$, we find 0.001 as a reasonable choice for CIFAR-10/100 and 0.0001 as a reasonable choice for ImageNet. For $\lambda$, we sweep the interval \{0.00001, 0.00005, 0.0001, 0.0005, 0.001, 0.005, 0.01\}. 
    Furthermore, we also consider three types of schedulers: {‘constant’, ‘linear’, ‘multistep(0.1-0.5:0.1)’} for $\lambda$ scheduling. Intuitively, a strong TrH regularization at the very beginning may lead the model to a flat highland instead of a flat minimum where we get a degenerated model. Ramping up $\lambda$ to the chosen value allows the model to focus more on accuracy and robustness at the early stage. In practice, we find that CIFAR10/100 is not sensitive to the choice of the $\lambda$ schedulers; however, ImageNet favors ‘linear’ or ‘multistep' schedulers over the 'constant' $\lambda$.
    \begin{itemize}
        \item ‘constant’. No $\lambda$ scheduling.
        \item ‘linear’. We ramp up $\lambda$ from 0 to the chosen value from iteration 1 to the end.
        \item ‘multistep(0.1-0.5:0.1)’. We use 0.01 * $\lambda$ before finishing 10\% of the total iterations. We use 0.1 * $\lambda$ after finishing 10\% and before 50\% of the total iterations. After finishing 50\% of iterations, we use $\lambda$. 
    \end{itemize}
\end{itemize}

\begin{table*}[t]
    \centering
    \small
    \scalebox{0.95}{\begin{tabular}{lllll}
     \multicolumn{5}{c}{\textbf{Fine-tuning  Hyper-parameters (CIFAR-10/100, ViT-L16)}}\\
    \toprule
    \toprule
     Defense& Hyper-parameter& Range & Final Choice & Final Choice  \\
     & &  &(CIFAR-10) & (CIFAR-100)  \\
     \midrule
     AT(\AWP) &$\delta_{awp}$ &\{0.0001, 0.0005, 0.001, 0.005, 0.01\} & 0.0005& 0.0005\\
     AT(\WA) & $\alpha$& - & 0.995 & 0.995\\
     AT(\STO) & $\alpha$ & - & 0.1 & 0.1 \\
     AT(\texttt{TrH}) & $\lambda$&\{0.00001, 0.00005, 0.0001, 0.0005, 0.001, 0.005, 0.01\} & 0.00001& 0.01\\
     &$\lambda$ schedule &\{`constant', `linear', `multistep'\} & `multistep' & `multistep'\\
     &$\gamma$ & -& 0.001&0.001\\
     TRADES(\AWP), $\lambda_t$=6 & $\delta_{awp}$&\{0.0001, 0.0005, 0.001, 0.005, 0.01\} & 0.0005 & 0.0001\\
     TRADES(\WA), $\lambda_t$=6& $\alpha$& - & 0.995 &0.995\\
     TRADES(\STO), $\lambda_t$=6& $\alpha$ & - & 0.3  & 0.3\\
     TRADES(\texttt{TrH}), $\lambda_t$=6 & $\lambda$&\{0.00001, 0.00005, 0.0001, 0.0005, 0.001, 0.005, 0.01\} & 0.0001 & 0.0001\\
     &$\lambda$ schedule & \{`constant', `linear', `multistep'\}& `multistep' & `constant'\\
     &$\gamma$ & - &  0.001& 0.001\\
         \bottomrule\\
    \end{tabular}}

    \scalebox{0.95}{\begin{tabular}{lllll}
     \multicolumn{5}{c}{\textbf{Fine-tuning  Hyper-parameters (CIFAR-10/100, Hybrid-L16)}}\\
    \toprule
    \toprule
     Defense& Hyper-parameter& Range & Final Choice & Final Choice  \\
     & &  &(CIFAR-10) & (CIFAR-100)  \\
     \midrule
     AT(\AWP) &$\delta_{awp}$ &\{0.0001, 0.0005, 0.001, 0.005, 0.01\} &0.0005 & 0.0005\\
     AT(\WA) & $\alpha$& - & 0.995 & 0.995\\
     AT(\STO) & $\alpha$ & - & 0.1 & 0.1 \\
     AT(\texttt{TrH}) & $\lambda$&\{0.00001, 0.00005, 0.0001, 0.0005, 0.001, 0.005, 0.01\} &0.0005& 0.0005\\
     &$\lambda$ schedule &\{`constant', `linear', `multistep'\} &`multistep' & `multistep'\\
     &$\gamma$ & -& 0.001& 0.001\\
     TRADES(\AWP), $\lambda_t$=6 & $\delta_{awp}$&\{0.0001, 0.0005, 0.001, 0.005, 0.01\} & 0.0005  & 0.0005\\
     TRADES(\WA), $\lambda_t$=6& $\alpha$& - & 0.995 &0.995\\
     TRADES(\STO), $\lambda_t$=6& $\alpha$ & - & 0.3  & 0.3\\
     TRADES(\texttt{TrH}), $\lambda_t$=6 & $\lambda$&\{0.00001, 0.00005, 0.0001, 0.0005, 0.001, 0.005, 0.01\} & 0.0001 & 0.0001\\
     &$\lambda$ schedule & \{`constant', `linear', `multistep'\}& `multistep' & `multistep'\\
     &$\gamma$ & - &0.001 &0.001 \\
         \bottomrule\\
    \end{tabular}}
    
    \caption{Hyper-parameters used in each defense and the values used to reproduce CIFAR10/100 results in Table.~\ref{tab:main_results}.}
    \label{tab:fine-tuning for cifar10/100}
\end{table*}

\begin{table*}[t]
    \centering
    \small
  \scalebox{0.95}{\begin{tabular}{lllll}
     \multicolumn{5}{c}{\textbf{Fine-tuning  Hyper-parameters (ImageNet, ViT-B16)}}\\
    \toprule
    \toprule
     Defense& Hyper-parameter& Range & Final Choice ($\ell_\infty$) & Final Choice ($\ell_2$) \\
     \midrule
     AT(\AWP) &$\delta_{awp}$ &\{0.0001, 0.0005, 0.001, 0.005, 0.01\} & 0.0001& 0.0001\\
     AT(\WA) & $\alpha$& - & 0.995 & 0.995\\
     AT(\STO) & $\alpha$ & - & 0.1 & 0.1 \\
     AT(\texttt{TrH}) & $\lambda$&\{0.00001, 0.00005, 0.0001, 0.0005, 0.001, 0.005, 0.01\} & 0.0001& 0.00005\\
     &$\lambda$ schedule &\{`constant', `linear', `multistep'\} & `multistep' & `linear'\\
     &$\gamma$ & -& 0.0001 &0.0001 \\
     TRADES(\AWP), $\lambda_t$=6 & $\delta_{awp}$&\{0.0001, 0.0005, 0.001, 0.005, 0.01\} & 0.0001 & 0.0001\\
     TRADES(\WA), $\lambda_t$=6& $\alpha$& - & 0.995 &0.995\\
     TRADES(\STO), $\lambda_t$=6& $\alpha$ & - & 0.3  & 0.3\\
     TRADES(\texttt{TrH}), $\lambda_t$=6 & $\lambda$&\{0.00001, 0.00005, 0.0001, 0.0005, 0.001, 0.005, 0.01\} & 0.00005 & 0.00005\\
     &$\lambda$ schedule & \{`constant', `linear', `multistep'\}& `linear'  & `linear'\\
     &$\gamma$ & - & 0.0001 &0.0001 \\
         \bottomrule\\
    \end{tabular}}

    \scalebox{0.95}{\begin{tabular}{lllll}
     \multicolumn{5}{c}{\textbf{Fine-tuning  Hyper-parameters (ImageNet, ViT-L16)}}\\
    \toprule
    \toprule
     Defense& Hyper-parameter& Range & Final Choice ($\ell_\infty$) & Final Choice ($\ell_2$) \\
     \midrule
     AT(\AWP) &$\delta_{awp}$ &\{0.0001, 0.0005, 0.001, 0.005, 0.01\} & 0.0001&0.0001\\
     AT(\WA) & $\alpha$& - & 0.995 & 0.995\\
     AT(\STO) & $\alpha$ & - & 0.1 & 0.1 \\
     AT(\texttt{TrH}) & $\lambda$&\{0.00001, 0.00005, 0.0001, 0.0005, 0.001, 0.005, 0.01\} & 0.0005&0.0005\\
     &$\lambda$ schedule &\{`constant', `linear', `multistep'\} &`multistep' &`linear' \\
     &$\gamma$ & -& 0.0001 &0.0001 \\
     TRADES(\AWP), $\lambda_t$=6 & $\delta_{awp}$&\{0.0001, 0.0005, 0.001, 0.005, 0.01\} & 0.0001 & 0.0001\\
     TRADES(\WA), $\lambda_t$=6& $\alpha$& - & 0.995 &0.995\\
     TRADES(\STO), $\lambda_t$=6& $\alpha$ & - & 0.3  & 0.3\\
     TRADES(\texttt{TrH}), $\lambda_t$=6 & $\lambda$&\{0.00001, 0.00005, 0.0001, 0.0005, 0.001, 0.005, 0.01\}& 0.0005 & 0.00005\\
     &$\lambda$ schedule & \{`constant', `linear', `multistep'\}& `multistep'& `multistep'\\
     &$\gamma$ & - & 0.0001 &0.0001 \\
         \bottomrule\\
    \end{tabular}}
    
    \caption{Hyper-parameters used in each defense and the values used to reproduce ImageNet results in Table.~\ref{tab:main_results}.}
    \label{tab:fine-tuning for imagenet}
\end{table*}



\section{Additional Results}
\begin{table*}[!t]
\centering
\begin{tabular}{lcccc}
\\
\toprule
\toprule
$\ell_\infty (\delta=8/255)$  & \multicolumn{4}{c}{ViT-B16}                                  \\ 
 \footnotesize{SE}$=\se{0.5}$\% &
  \multicolumn{2}{c}{CIFAR-10} &
  \multicolumn{2}{c}{CIFAR-100} \\ 
Defense &
  \texttt{Clean}(\%) &
  \texttt{AA}(\%) &
  \texttt{Clean} (\%)&
  \texttt{AA}(\%) \\ \midrule
AT(\base)     & 87.5 & \multicolumn{1}{c|}{60.3} & 60.0 & 30.4 \\
AT(\WA)       &  86.9& \multicolumn{1}{c|}{60.4} &64.1 & \textbf{33.7} \\
AT(\STO)   &86.8  & \multicolumn{1}{c|}{60.4} & 63.2 &31.5  \\
AT(\AWP)  & 87.3  & \multicolumn{1}{c|}{\underline{61.3}} & 61.5 & 32.2 \\
AT(\texttt{TrH})     & 88.4 & \multicolumn{1}{c|}{\textbf{61.5}} & 65.0 & \underline{33.0} \\ \midrule
TRADES(\base)  & 85.4  & \multicolumn{1}{c|}{\underline{60.8}} & 58.6 & 30.5 \\
TRADES(\WA)   & 85.6 & \multicolumn{1}{c|}{\underline{60.6}} & 62.7 & \textbf{33.0} \\
TRADES(\STO) & 85.9 & \multicolumn{1}{c|}{\textbf{61.1}} & 63.5 & 31.5 \\
TRADES(\AWP)   &  84.7& \multicolumn{1}{c|}{\underline{60.1}} & 60.8 & \underline{32.2} \\

TRADES(\texttt{TrH})  & 85.8 & \multicolumn{1}{c|}{\textbf{61.1}} & 63.8  & \textbf{33.0} \\ 

\bottomrule
\end{tabular}

\caption{Additional results for CIFAR-10/100 using ViT-B16. \texttt{Clean}: \% of Top-1 correct predictions. \texttt{AA}: \% of Top-1 correct predictions under AutoAttack. A max Standard Error (SE)~\cite{stark2005sticigui} = $\sqrt{0.5*(1-0.5)/m}$ ($m$ as the number of test examples) is computed for each dataset. The best results appear in bold. Underlined results are those that fall within the SE range of the result and are regarded roughly equal to the best result.}
\label{tab:additional_cifar_results}
\end{table*}

\begin{table*}[!t]
\centering
\begin{tabular}{lcccc}
\\
\toprule
\toprule
ImageNet  & \multicolumn{4}{c}{Hybrid-L16}                                  \\ 
 \footnotesize{SE}$=\se{0.2}$\% &
  \multicolumn{2}{c}{$\ell_\infty (\delta=4/255)$} &
  \multicolumn{2}{c}{$\ell_2 (\delta=3.0)$} \\ 
Defense &
  \texttt{Clean}(\%) &
  \texttt{AA}(\%) &
  \texttt{Clean} (\%)&
  \texttt{AA}(\%) \\ \midrule
AT(\base)     & 72.6 & \multicolumn{1}{c|}{40.7} & 72.2 & 40.6 \\
AT(\WA)       & 72.7 & \multicolumn{1}{c|}{40.4} & 72.7 &  40.5\\
AT(\STO)   & 72.8 & \multicolumn{1}{c|}{43.6} & 72.3 & 40.9 \\
AT(\AWP)     & 67.7 & \multicolumn{1}{c|}{39.4} & 67.9 & 40.3 \\
AT(\texttt{TrH})     & 75.0 & \multicolumn{1}{c|}{\textbf{46.2}} & 74.4 &  \textbf{45.9}\\ \midrule
TRADES(\base)  & 79.5 & \multicolumn{1}{c|}{37.6} &  78.0& 38.6 \\
TRADES(\WA)   & 72.9 & \multicolumn{1}{c|}{40.9} & 71.3 & 40.8 \\
TRADES(\STO) & 73.8 & \multicolumn{1}{c|}{41.3} & 72.2 & 41.2 \\
TRADES(\AWP)   & 66.4 & \multicolumn{1}{c|}{38.8} & 65.3 & 40.9 \\

TRADES(\texttt{TrH})  & 75.9 & \multicolumn{1}{c|}{\textbf{45.7}} & 73.3 & \textbf{45.6} \\ 

\bottomrule
\end{tabular}

\caption{Addtional Results for ImageNet using Hybird-L16. \texttt{Clean}: \% of Top-1 correct predictions. \texttt{AA}: \% of Top-1 correct predictions under AutoAttack. A max Standard Error (SE)~\cite{stark2005sticigui} = $\sqrt{0.5*(1-0.5)/m}$ ($m$ as the number of test examples) is computed for each dataset. The best results appear in bold. Underlined results are those that fall within the SE range of the result and are regarded roughly equal to the best result.}
\label{tab:additional_imagenet_results}
\end{table*}

Due to space limitation, we include the results on CIFAR-10/100 with ViT-B16 in Table~\ref{tab:additional_cifar_results} and ImageNet with Hybrid-L16 in Table~\ref{tab:additional_imagenet_results} of the appendix. These additional results are consistent with what we have found in Section~\ref{sec:evaluation-result} of the paper: in CIFAR-10/100, TrH is consistently among the top or silver results; in ImageNet, TrH has significantly advantages over the other baseline methods.

\section{ViT Architecture}\label{appendix:architecture}
We describe the architectures of Vision Transformers used in our experiments in Section~\ref{sec:evaluation}. 
\begin{itemize}
    \item ViT-B16 includes 12 transformer layers with hidden size 768, MLP layer size 3072 and 12 heads in the multi-head attention. 
    \item ViT-L16 includes 24 transformer layers with hidden size 1024, MLP layer size 4096 and 16 heads in the multi-head attention. 
    \item Hybird-L16 is a hybrid model of a ResNet-50 and a ViT-L16 model. The patches fed into ViT are feature representations encoded by a ResNet-50 and projected to the Transformer dimensions~\cite{dosovitskiy2021an}. Dissimilar to a standard ResNet-50 feature encoder, Dosovitskiy et al.~\cite{dosovitskiy2021an} replaced Batch Normalization with Group Normalization~\cite{Wu2019GroupN} and use standardized Convolution~\cite{qiao2019micro}. Moreover, Dosovitskiy et al.~\cite{dosovitskiy2021an} removed stage 4, placed the same number of layers in stage 3 (keeping the total number of layers), and took the output of this extended stage 3 as the input of ViT. 
\end{itemize}

The pretrained ViT-B16 and ViT-L16 checkpoints use patch size $16\times 16$. However, since the images in CIFAR10/100 are of size $32\times 32$, there will be only 4 patches when using the original patch size. Therefore, we downsample the kernel of the first convolution to produce image patches of $4 \times 4$. Among different ways of downsampling we find the \texttt{cubic} interpolation gives the best results, which is the one chosen in pre-tuning as discussed in Appendix~\ref{appendix:pre-tuning}. 
Indeed, Mahmood et al.~\cite{Mahmood_2021_ICCV} empirically finds that such down-sampling provides better results for CIFAR10 images. Furthermore, in generating patches, we also use a stride of 2 instead of 4, because we find using overlapped patches increases the sequence length and results in better performance.  

\section{TrH Regularization for Other Adversarial Losses}\label{appendix:other-loss}

In this section, we apply TrH regularization to some additional robust losses other than AT or TRADES.

\subsection{ALP~\cite{Kannan2018AdversarialLP}}
Similar to TRADES, Adversarial Logit Pairing (ALP)~\cite{Kannan2018AdversarialLP} is another method that pushes points away from the decision boundary by regularizing the $\ell_2$ difference between the clean and the adversarial softmax outputs. Formally, ALP minimizes the following loss during training,
\begin{align*}
    \hat{R}_{\texttt{A}}(\theta, D^m) &\defeq \frac{1}{m}\sum_{(x, y)\in D^m} \CE((x', y), g_\theta) + \lambda_{A}||s(g_\theta(x)) - s(g_\theta(x'))||^2_2 \\
    \text{ where } x' &= x + \arg\max_{||\epsilon||_p \leq \delta} \CE((x+\epsilon, y), g_\theta).
\end{align*}

We hereby derive TrH regularization for $\hat{R}_{\texttt{A}}(\theta, D^m) $. Using Proposition~\ref{prop:at_trh}, we know that 
\begin{align*}
    \TrH(\CE((x', y), g_\theta)) = ||f_{\theta_b}(x')||^2_2 \cdot \mathbf{1}^\top h(x', \theta).
\end{align*}
The rest of this section will focus on the $\ell_2$ distance loss, 
\begin{align}
    S \defeq ||s(g_\theta(x)) - s(g_\theta(x'))||^2_2 = \sum_i (s(g_i) - s(g'_i))^2,
\end{align} where $g = g_\theta(x), g' = g_\theta(x')$. To compute $\Tr( \triangledown^2_{\theta_t}S)$, we need to re-use Lemma~\ref{lemma:softmax_grad} and~\ref{lemma:log_softmax_grad} to obtain the derivatives of the softmax and log-softmax outputs, i.e. 
\begin{align*}
    \frac{\partial s(g_\theta(x))}{g_\theta(x)} &= \Jprob \defeq diag(s(g_\theta(x))) - s(g_\theta(x))\cdot s(g_\theta(x))^\top\\
    \text{ and } \;\; \frac{\partial \log s(g_\theta(x))}{g_\theta(x)} &= \Jlprob \defeq I - \mathbf{1}\cdot s(g_\theta(x))^\top,
\end{align*}
as well as
\begin{align*}
    \frac{\partial \Jprob_{ik}}{\partial \{g_{\theta}(x)\}_k} = \Jprob_{kk}\Jlprob_{ki}.
\end{align*}
For the ease of notation, we let $w \defeq \theta_t$ be the weights of the top-layer. Now we are ready to write the first-order derivative of $S$ w.r.t $w_{jk}$ as follows
\begin{align*}
    \frac{\partial S}{\partial w_{jk}} &= \sum_i 2(s(g_i) - s(g'_i))\left[ \frac{\partial s(g_i)}{\partial w_{jk}} - \frac{\partial s(g'_i)}{\partial w_{jk}}\right]\\
    &=2\sum_i (s(g_i) - s(g'_i))(\Jprob_{ik}\{f_{\theta_b}(x)\}_j - \Jprob'_{ik}\{f_{\theta_b}(x')\}_j).
\end{align*}
The second-order derivative is equal to
\begin{align}
    \frac{\partial^2 S}{\partial w^2_{jk}} &= 2\sum_i\left[ \frac{\partial s(g_i)}{\partial w_{jk}} - \frac{\partial s(g'_i)}{\partial w_{jk}}\right](\Jprob_{ik}\{f_{\theta_b}(x)\}_j - \Jprob'_{ik}\{f_{\theta_b}(x')\}_j) + (s(g_i) - s(g'_i))\left[ \frac{\partial \Jprob_{ik}}{\partial w_{jk}}\{f_{\theta_b}(x)\}_j - \frac{\partial \Jprob'_{ik}}{\partial w_{jk}}\{f_{\theta_b}(x')\}_j\right] \nonumber\\
    &=2\sum_i (\Jprob_{ik}\{f_{\theta_b}(x)\}_j - \Jprob'_{ik}\{f_{\theta_b}(x')\}_j)^2 + (s(g_i) - s(g'_i))(\Jprob_{kk}\Jlprob_{ki}\{f_{\theta_b}(x)\}^2_j - \Jprob'_{kk}\Jlprob'_{ki}\{f_{\theta_b}(x')\}^2_j).\label{eq:ALP-second-order-derivative}
\end{align}
Similar to TRADES, if one stops gradients over the clean logit $g$, then Eq.~\ref{eq:ALP-second-order-derivative} can be simplified as,
\begin{align*}
    \frac{\partial^2 S}{\partial w^2_{jk}} &= 2\sum_i (\Jprob'_{ik}\{f_{\theta_b}(x')\}_j)^2 + (s(g_i) - s(g'_i))(- \Jprob'_{kk}\Jlprob'_{ki}\{f_{\theta_b}(x')\}^2_j)\\
    &= 2\{f_{\theta_b}(x')\}^2_j \Big(||\Jprob_{k}||^2_2 - \Jprob'_{kk} ((s(g) - s(g'))^\top\Jlprob'_k )  \Big).
\end{align*}
Therefore, the trace of Hessian is equal to,
\begin{align*}
\Tr(\triangledown^2_{\theta_t}S) &= \sum_{jk} 2\{f_{\theta_b}(x')\}^2_j \Big(||\Jprob_{k}||^2_2 - \Jprob'_{kk} ((s(g) - s(g'))^\top\Jlprob'_k )\\
&=2 ||f_{\theta_b}(x')||^2_2 \cdot \sum_k \Big(||\Jprob_{k}||^2_2 - \Jprob'_{kk} ((s(g) - s(g'))^\top\Jlprob'_k )  \Big).
\end{align*}
If the gradient over $g$ is kept, then one can sum over the feature dimension $j$ and the class dimension $k$ in Eq.~\ref{eq:ALP-second-order-derivative}. To put together, the TrH regularization term for ALP is given as follows:
\begin{align*}
    \frac{1}{m}\sum_{(x, y)\in D^m} ||f_{\theta_b}(x')||^2_2 \cdot \mathbf{1}^\top h(x', \theta) + \lambda_A \cdot\Tr(\triangledown^2_{\theta_t}S).
\end{align*}

\subsection{MART~\cite{Wang2020Improving}}

In~\cite{Wang2020Improving}, Wang et al. proposed MART as a robust training loss that focus more on points that are not classified correctly. Different from AT, TRADES and ALP, MART explicitly aims to increase the margin between the top prediction and the second best candidate. 
In what follows we provide the TrH regularization for MART. 
We denote the MART loss as $\hat{R}_{\texttt{M}}$, which contains two components: a boosted Cross-Entropy (BCE) loss and a weighted KL-Divergence (WKL),
\begin{align*}
    \hat{R}_{\texttt{M}}(\theta, D^m) \defeq \frac{1}{m} \sum_{(x, y)\in D^m} \BCE((x', y), g_\theta) + \lambda_m \WKL((x, x'), g_\theta), 
\end{align*}where
\begin{align*}
    \BCE((x', y), g_\theta) &\defeq \CE((x, y), g_\theta) + (-\log(1-\max_{\kappa \neq y}s(\{g_\theta(x')\}_\kappa)),\\
    \text{ and, }\; \WKL((x, x'), g_\theta) &\defeq  \KL(s(g_\theta(x))||s(g_\theta(x')))(1-s(\{g_\theta(x))\}_y).
\end{align*}
Here
\begin{align}
    x' = x + \arg\max_{||\epsilon||_p \leq \delta} \CE((x, y), \theta).
\end{align}
First, we derive TrH of the BCE loss with respect to the top-layer weights $\theta_t$,
\begin{align*}
\Tr(\triangledown^2_{\theta_t}\BCE)=\Tr(\triangledown^2_{\theta_t} [\CE((x, y), g_\theta)]) + \Tr( \triangledown^2_{\theta_t}[-\log(1-\max_{\kappa \neq y}s(\{g_\theta(x')\}_\kappa)]).
\end{align*}
Using intermediate steps from the proof of  Proposition~\ref{prop:trades_trh} in Appendix~\ref{appendix:proof}, we have that 
\begin{align*}
\Tr(\triangledown^2_{\theta_t} [\CE((x, y), g_\theta)]) = ||f_{\theta_b}(x)||^2_2 \cdot \mathbf{1}^\top h(x, \theta).
\end{align*}
In order to compute $\Tr( \triangledown^2_{\theta_t}[-\log(1-\max_{\kappa \neq y}s(\{g_\theta(x')\}_\kappa)])$, we make use of Lemma~\ref{lemma:softmax_grad} and~\ref{lemma:log_softmax_grad} to obtain the derivatives of the softmax and log-softmax outputs,
\begin{align*}
    \frac{\partial s(g_\theta(x'))}{g_\theta(x')} &= \Jprob' \defeq diag(s(g_\theta(x'))) - s(g_\theta(x'))\cdot s(g_\theta(x'))^\top\\
    \text{ and } \;\; \frac{\partial \log s(g_\theta(x'))}{g_\theta(x')} &= \Jlprob' \defeq I - \mathbf{1}\cdot s(g_\theta(x'))^\top,
\end{align*}
as well as,
\begin{align*}
    \frac{\partial \Jprob'_{ik}}{\partial \{g_{\theta}(x')\}_k} = \Jprob'_{kk}\Jlprob'_{ki}.
\end{align*}
Let us denote $K=-\log(1-\max_{\kappa \neq y}s(\{g_\theta(x')\}_\kappa)]$ and $w = \theta_t$ for the ease of notation. Then the first-order derivative of $K$ w.r.t. $w_{jk}$ is,
\begin{align*}
    \frac{\partial K}{\partial w_{jk}}= \frac{\Jprob'_{\kappa^*k} }{1-s(\{g_\theta(x')\}_{\kappa^*})}\{f_{\theta_b}(x')\}_j,
\end{align*} where
\begin{align*}
    \kappa^* = \arg\max_{\kappa \neq y}s(\{g_\theta(x')\}_\kappa).
\end{align*}
The second-order derivative is,
\begin{align*}
\frac{\partial^2 K}{\partial w^2_{jk}} &= \frac{\frac{\partial \Jprob'_{\kappa^* k}}{\partial w_{jk}}(1-s(\{g_\theta(x')\}_{\kappa^*})) +\Jprob'_{\kappa^* k}\Jprob'_{\kappa^* k}\{f_{\theta_b}(x')\}_j  }{(1-s(\{g_\theta(x')\}_{\kappa^*}))^2}\{f_{\theta_b}(x')\}_k\\
&=\frac{\Jprob'_{kk}\Jlprob'_{k\kappa^*}\{f_{\theta_b}(x')\}_k(1-s(\{g_\theta(x')\}_{\kappa^*})) + \Jprob'_{\kappa^* k}\Jprob'_{\kappa^* k}\{f_{\theta_b}(x')\}_j}{(1-s(\{g_\theta(x')\}_{\kappa^*}))^2}\{f_{\theta_b}(x')\}_j\\
&=\{f_{\theta_b}(x')\}^2_j \left[\frac{\Jprob'_{\kappa^* k}\Jprob'_{\kappa^* k}}{1-s(\{g_\theta(x')\}_{\kappa^*})} +\frac{\Jprob^{'2}_{\kappa^* k}}{(1-s(\{g_\theta(x')\}_{\kappa^*}))^2}\right].
\end{align*}
Thus, 
\begin{align*}
    \Tr(\frac{\partial^2 K}{\partial w^2_{jk}}) &= \sum_{jk} \{f_{\theta_b}(x')\}^2_j \left[\frac{\Jprob'_{\kappa^* k}\Jprob'_{\kappa^* k}}{1-s(\{g_\theta(x')\}_{\kappa^*})} +\frac{\Jprob^{'2}_{\kappa^* k}}{(1-s(\{g_\theta(x')\}_{\kappa^*}))^2}\right]\\
    &= ||f_{\theta_b}(x')||^2_2 \sum_k \left[\frac{\Jprob'_{\kappa^* k}\Jprob'_{\kappa^* k}}{1-s(\{g_\theta(x')\}_{\kappa^*})} +\frac{\Jprob^{'2}_{\kappa^* k}}{(1-s(\{g_\theta(x')\}_{\kappa^*}))^2}\right],
\end{align*}
and 
\begin{align*}
\Tr(\triangledown^2_{\theta_t}\BCE) = ||f_{\theta_b}(x)||^2_2 \cdot \mathbf{1}^\top h(x, \theta) + ||f_{\theta_b}(x')||^2_2 \sum_k \left[\frac{\Jprob'_{\kappa^* k}\Jprob'_{\kappa^* k}}{1-s(\{g_\theta(x')\}_{\kappa^*})} +\frac{\Jprob^{'2}_{\kappa^* k}}{(1-s(\{g_\theta(x')\}_{\kappa^*}))^2}\right].
\end{align*}

Next, we derive the trace of Hessian for WKL loss. Similarly to before, by stopping the gradient over the clean logit $g_\theta(x)$, the trace of Hessian of WKL will be the same as the KL loss used in TRADES, which has been shown in Proposition~\ref{prop:trades_trh}. 
Namely, in this case
\begin{align*}    \Tr(\triangledown^2_{\theta_t}\WKL) =  ||f_{\theta_b}(x')||^2_2 \cdot \mathbf{1}^\top h(x', \theta).
\end{align*} 
On the other hand, if the gradient on $g_\theta(x)$ is computed, there is an additional term in $\Tr(\triangledown^2_{\theta_t}\WKL)$ (similar but more complicated than $G(x, x';\theta)$ derived in the proof of Proposition~\ref{prop:trades_trh} in Appendix~\ref{appendix:proof}), which we will leave as future work. Finally, we present the TrH regularization for MART as follows,
\begin{align*}
    \TrH_{\texttt{M}}(x, x';\lambda_m, \theta) = \Tr(\triangledown^2_{\theta_t}\BCE) + \lambda_m \Tr(\triangledown^2_{\theta_t}\WKL).
\end{align*}
This concludes our derivations of the TrH regularization for the MART loss.

\end{appendices}

\end{document}